# A FORTRAN coded Regular Expression Compiler for the IBM 1130 Computing System[†]

Gerardo Cisneros[*]

**ABSTRACT,**

REC (Regular Expression Compiler) is a concise programming language which allows students to write programs without knowledge of the complicated syntax of languages like FORTRAN and ALGOL. The language is recursive and contains only four elements for control. This paper describes an interpreter of REC written in FORTRAN.

*Acknowledgement*

The programs forming, the REC system were written for tile IBM 1130 computer of the Centro Nacional de Calculo of the Instituto Politécnico Nacional, which generously provided all the necessary facilities.

INTRODUCTION

Programming has become an important part of introductory Numerical Analysis courses. REC, with its simple form of control offers the student a reduction in the time spent programming and debugging as well as in learning the language itself. The processor described provides the student with all tile arithmetic operations and functions supplied by FORTRAN; fixed numerical output format and free input format allow him to concentrate on more import ant aspects of programming.

REC is based on the concept of a regular expression, wish arises in automata theory. A rigorous definition of the language was published by H. V. McIntosh in the Acta Mexicana de Ciencia y Tecnología. For our present purposes it is only necessary to know that a regular expression over an alphabet is defined as a finite string of letters, of the alphabet, composed following the rules for concatenation, iteration and, selection of alternatives. We will treat the letters as the names of arithmetic or other operations which we are interested in executing. Concatenation then means that we wish perform several operations in sequence, iteration means that we wish to repeat a sequence, and of course it is necessary to distinguish alternative courses of computation in order to know when to terminate a given course of action, or to adapt one of several possible courses to the situation at hand.

---

[†] This version of REC is archaeological reconstruction of REC/A language on IBM1130 Simulator (SIMH IBM 1130 Emulator and Disk Monitor System R2V12) from Computer History Simulation Project (www.ibm1130.org), also see "REC language is a live" for Ignacio Vega-Páez

[*] Escuela Superior de Física y Matemáticas, Instituto Politécnico Nacional, México.

To specify a programming language in terms of regular expressions, delimiters, separators and predicates are introduced. Every letter of the alphabet (which does not contain the delimiters or separators) is a predicate and furthermore it represents an operation. A predicate is a function which may have one of two truth values as a result of the execution of the operation in represents. These values shall be called *true* and *false*. Some predicates are always *true* and are called operators. Delimiters are represented by parentheses and are used denote a single expression, which should itself be treated as a composite predicate; colon and semicolon are separators.

Concatenation is implied by writing expressions in sequence, colon implies iteration of the expression starting at the left delimiter of the corresponding level and semicolon terminates the concatenation of a string. Since concatenation is associative, REC expressions should be written in an extended form without parentheses.

Reading from left to right, operations are executed in the order encountered, taking into account the truth value of each letter scanned. If the value is *true*, the next operation in sequence is executed. Should the truth value be *false*, control goes immediately to the next separator or right parenthesis (of the expression in which the letter appears) and execution continues with the first letter that follows. If the bounding right parenthesis of the current expression is encountered in normal course of operation, execution of the expression is complete and its truth value is *false*, causing a search for the next separator or right parenthesis.

If a colon is found, execution of the expression repeats from the beginning; if a semicolon is encountered, execution of the expression is complete and it has the truth value *true*, thus resuming execution following the corresponding right delimiter.

It is observed from this that the only elements of control are the parentheses, colon and semicolon.

In order that the beginning and ending of a program are unambiguous, opening and closing parentheses are needed. Once a program has been written this way, it may be declared as a predicate by assigning it a new letter. It may then be used as a subroutine by simply writing the corresponding letter.

Inasmuch as it is defined quite abstractly, REC is independent of a particular computer. The version described in this article is written in FORTRAN so that it may be used on as wide a class of computers as possible, and so that its details will refer to a programming language which is widely understood and generally available. Nevertheless there are many points which depend upon the IBM 1130 FORTRAN, particularly the binary representation of Hollerith characters. We assume that the reader is familiar with FORTRAN for the 1130 System.

The programs comprising the REC system fall into several convenient groupings. First we have subroutines handling the input output and the conversion between binary numbers and EBCDIC character strings. Similar conversion programs may be found among the IBM system subroutines, but have been reproduced here exclusively in FORTRAN to maintain full machine independence and greater compatibility with the remainder of the REC processor. These routines are:

| | | |
|---|---|---|
| RECR | Input of characters through a buffer. | |
| RECW | Output of characters through a buffer.* | |
| RECNE | Reduction of EBCDIC symbols to the range 1-64. | |
| RECNC | Conversion of character strings to floating point or integer binary numbers. | |
| RECNP | Conversion of a floating point number to an EBCDIC character string | |

The cycle of operation of the REC system falls into three distinct phases the monitor phase, the compilation phase, and the execution phase. Each phase has its own collection of subroutines, responsible for specific aspects of the phase.

Monitor Phase:
    RECMO   Contains all of the Monitor.

Compiler Phase:
    RECCM   Controller for the phase.
    RECCO   Compiles the control characters
    RECCA   Compiles predicates or operators, including those which use numerical or EBCDIC arguments.
    RECCC   Compiles program constants.
    RECKO   Compiles counters.
    RECQU   Compiles quoted character strings.
    RECFC   Fills exit chains formed during the compilation.

Execution Phase:

    RECXC   Controls the execution phase.
    RECXQ   Contains the definition of the specific operators characterizing REC/A (REC/Arithmetic)
    RECCN   Executes counters.
    RECDS   Tests the data switches, allowing program interruption.

Finally there is the main program driving REC, as well as two service programs which load the disk with the directories and error messages --REC, RECDO and RECER, respectively.

Each of these programs will be described in detail. In the appendix an annotated listing of them is included, arranged according to their alphabetical order.

Most of the variables used in the program are carried in COMMON, so that the subroutines can use a minimum of arguments. Each variable has its own use, which is uniform throughout all the subroutines.

The expressions compiled by the program described in this paper perform arithmetic calculations. All the operations and functions supplied by FORTRAN are available as predicates or operators. REC programs are written in reversed polish notation, making the compilation of expressions a very simple process.

*I. Variables in COMMON*

The assignment of the variables in COMMON is:

| | |
|---|---|
| IAC | Integer Accumulator. This variable is used to transmit EBCDIC characters or special REC character codes. |
| ILC | Instruction Location Counter. Points to the next available cell in IPROG during compilation. |
| IXL | Execution Location Counter. Pointer on IPROG during execution. |
| ILC0 | Beginning of the program being compiled. |
| IR | Input buffer pointer. |
| IW | |
| IX | Variable used in computed GO TO indexing. |
| IP | Compilation push-down list pointer. |
| IE | Error flag |
| IIN | Current input device: IIN = 2 indicates card reader, =6 means console keyboard. |
| IOUT | Current output device. IOUT equals 1 for typewritten output, 2 for punched output and 3 for printed output. These device numbers are consistent with 1130 FORTRAN device numbers. |
| IL | Echo suppression flag, |
| IPROG | REC programs are compiled into this array of length 500. |
| IPDL | Push-down list used during compilation, of dimension (10,3). |
| ICPL | Contains the compilation code for each character. It can hold up to 128 definitions. |
| IXEQ | Contains execution codes. Like ICPL, IXEQ has a length of 128 words. |
| IM | Pointer on the arithmetic pushdown list (c. f. Sec. 3). |
| KS | Main pointer for the program constant array. |
| KS1 | Auxiliary pointer on CONST. |
| CONST | Program constant storage. Up to 30 floating point constants can be stored in this array. |

*2. Input, Output and Conversion subprograms.*

    i) RECR.

This subroutine tests IR when entered. If IR is greater than 80, it reads a record from device IIN into ICARD, sets IR to 1 and places ICARD(IR) in IAC. IR is then incremented. If IR is not greater than 80, RÉCR places ICARD(IR) in IAC and increments IR. It converts the IBM 029 punches for percent, lozenge, number sign and at sign (026 punches (,) = and at sign) into the Al format codes for (,) = and ' when input is done from cards. All non disk input is read by this routine

    ii) RECW.

RECW is used for character and error message output. If the calling argument J is 1, IW is increased by 1 and IAC is placed in ILINE(IW) or a return is executed, depending on whether IL is 1 or 2. If IW is equal to or greater than the allowed size of the output buffer (120 columns for the line printer, 80 for card punch and typewriter), ILINE is printed on the output device designated by IOUT and IW is set to zero. If the calling parameter is 2, ILINE is printed immediately without respect to whether it is completely full or not. If the buffer is empty nothing is done, but in any event IW is reset to zero so that the collection of a new line may begin. When RECW is called with J=3, it reads from disk storage and outputs on the current output unit the error message specified by the error flag, IE.

iii) RECNE.

In order to use characters to subscript the arrays ICPL, IXEQ and IMON (contained in RECMO), a derived code in the range 1-64 is needed. Since Al format characters are left justified, IAC is loaded into the accumulator, the two leftmost bits are shifted out and the resulting six bit number is right justified and stored in IAC. IAC is then incremented, since FORTRAN does not acknowledge zero subscripts. RECNE is the only subprogram written in assembly language.

iv) RECNC.

Depending on the calling parameter JJ, RECNC converts a character string of the form sx.xEsdd into a floating point number or a string in the form sx into a fixed point quantity. In these strings, s represents a sign which is optional, x is a string of zero or more digits, indicates an also optional decimal point and Esdd indicates an exponent of ten which may or may not appear, in which the d's are decimal digits. A floating point conversion stores its result in FCONV and a fixed point conversion output its value through ICONV. Any character that does not conform to the symbols which should appear in the part of the number which is being translated, like a second decimal point, a blank, etc., terminates the conversion and remains in IAC. Leading blanks are ignored.

If JJ is zero, RECNC converts a floating point number without writing each character as it is read, JJ = 1 implies also a floating point conversion, but with the writing of each character read and JJ = 2 is used for integer conversion, which is always done with individual character writing, since this kind of conversion is used only by the compilation phase, which outputs its results through RECW.

v) RECNP.

RECNP encodes its floating point argument into a string of EBCD coded characters in the form bsd.dddddEsdd and outputs it through RECW. The sign of the argument F is recorded in IS in the form of a blank character or a minus sign and the absolute value of F is placed in V which is then shifted into the range $1 = V < 10$; K counts the number of shifts and is thus used as the exponent of ten. V is rounded and the conversion takes place. If the output buffer pointer IW is greated than 107, RECW is called with an argument of two. This prevents the encoded string from being printed on two lines when output is done on the line printer.

3. *The main program and list of the available operators.*

    After the compiler has been loaded for execution by the 1130 Disk Monitor System, REC initializes the compiler and executor tables, sets the pointers to their initial values and calls the monitoring phase. On return, the compiler is called, and if this phase is successful, execution proceeds. Expressions do not compile into machine language, so execution requires an interpreter and a program containing internal operator definitions. This independence from machine language will result in slower execution on account of the interpreter, but has the advantage that REC can readily be transferred from one type of computer to another.

    REC defines the files used by the program. File 1 contains the necessary tables for the three phases and file 2 contains error messages.

    The expressions written in the REC language make use of a pushdown list, which is the type of list in which the most recently inserted element is the first one to be removed. Operators performing a calculation on the element (or elements) on top of the pushdown list replace their arguments with the evaluated function. There are arithmetic predicates that also use the pushdown list, but do not erase their arguments. Since their value is indicated internally through a skip, they do not deposit a value on the pushdown list.

    In addition to arithmetic predicates and operators there are also several others which, in conjunction with RECR and RECW, manipulate characters, so that comments or programmed commands may be included in a REC program.

    In the description to follow, the top of the pushdown list, which contains the most recently evaluated subexpression, will be called the "accumulator." Since many operators use two arguments, we will call the second element on the list the "first operand." The rationale of this nomenclature is that we may regard expressions such as abcd+++ as the repetition of a monary operator in which a, b, c, d in turn are "accumulated" into a single sum.

| | |
|---|---|
| A | Computes the absolute value of the quantity in the accumulator. |
| B | Raises the first operand in the list to the power given by the accumulator. |
| C | Computes the cosine of the number in the accumulator. |
| E | Calculates the exponential (to the base e) of the quantity in the accumulator. |
| Fk | Fetches the value stored at the variable k (k is a digit from 0 to 9) and pushes it onto the list. |
| H | Evaluates the hyperbolic tangent of the quantity in the accumulator. |
| I | Inputs a number and pushes it onto the list. the datum is indicated in the input record by the characters '/ and is finished by another apostrophe; blanks between these delimiters and the quantity read are ignored. Only blanks are skipped when searching for the combination '/, other symbols cause an error message to be printed. Data should start after the third column following the closing right parenthesis   (c. f. RECCO, Sec. 5). |
| J | Tests the accumulator and the first operand in the list for equality. It is predicate that takes the *true* exit if the numbers compared are equal within the range of 0.000005. |
| L | Lifts the pushdown list, i.e., it removes the element on the top of the list. |
| M | Changes the sign of the contents of the accumulator. |
| N | Is a predicate that has the *truth* value true if the accumulator is negative. |
| O | Outputs the contents of the accumulator. The pushdown list remains intact. |
| P | Copies the accumulator onto the list, i.e., it repeats it. Thus, xP* results in x*x. |
| Q | Computes the square root of the number in the accumulator. |
| R | Reads the next character from the input buffer into IAC by use of RECR. The accumulator is not affected. |
| Sk | Stores the contents of the accumulator at the variable k. It leaves the pushdown list unaffected. |
| W | Writes IAC by use of RECW (the contents of IAC are sent to the output buffer). |
| X | Prints, types or punches the contents of the output buffer, without waiting, for a full line. It is then cleared, ready to begin a new line. |
| 'A | Computes the arctangent of the quantity in the accumulator. |
| 'L | Calculates the natural logarithm of the quantity in the accumulator. |
| 'S | Computes the sine of the number in the accumulator. |
| 0 | (Digit 0) Tests the accumulator for zero. It takes the value true if its contents are less in absolute value than 0.000005. |
| + or & | Add the contents of the accumulator to the value of the first operand in the list. |
| - | Subtracts the contents of the accumulator from the first operand in the list. Thus ab- is equivalent to a-b. |
| * | Computes the product of the accumulator and the first operand. |
| / | Divides the first operand in the list by the quantity in the accumulator. Hence ab/ is the reversed Polish of a/b. |
| '/ | Program constant operator. At execution time the constant is fetched from the location in CONST indicated by the argument in IPROG, pushing it onto the list, The format of a program constant is the same as for data outside the program. |
| " | Outputs through RECW the quoted string compiled by RECQU into IPROG. |
| =x | Compares the contents of IAC to the EBCDIC character x. It is a predicate; *true* if the character in IAC is the same as x. |
| $n$ | Defines a counter. A counter is a predicate with n+1 internal states. Each time it is encountered in the course of execution of a program in which it is contained, it advances from one state to the next, cyclically. It always yields the value *true*, except when it reaches the final state of the cycle, whereupon it yields the value *false* and prepares to repeat its cycle. Care should be taken in entering for a second time a program which contains a cycle which may not have run to completion. |

4. *Monitor Phase.*

Upon entrance to the REC Monitor (RECMO):

The monitor character codes are read from disk storage into the array IMON.
The error flag, IR, is reset to 1.
IR is set to 81 and RECR is called.

IAC, which contains the character in the first card column, is tested. If it contains a C, the rest of the record is treated as a commentary and this is put out without further testing. If IAC contains an asterisk, the monitor scans subsequent columns (and cards) for monitor control specifications until a left parenthesis is found, in which case the compilation phase is entered.

Six control specifications are defined. The definitions are equated to characters through IMON(64), which contains a value suitable for computed GO TO indexing for each of 64 characters. For technical reasons, an additional character is read and tested for left parenthesis when a control character is found. If. the specification uses an argument, it will be provided by the new symbol. The six specifications are:

| | |
|---|---|
| Ik | Define k as the input device (IIN = k). |
| Ok | Define k as the output device (IOUT = k). If k is not a valid device, an error message is issued. |
| T | Returns to the 1130 Disk Monitor System (performs a CALL EXIT). |
| E | Erases any existing REC defined subroutines from IPROG. |
| Nx.or N'x | Defines x or 'x as a recursive predicate, the program definition of which must be written in REC. |
| S | Suppresses character echoing, and is used when a program listing is not desired, be it for reasons of timing or of familiarity. |

RECMO calls all the I/O and conversion subroutines but RECNP.

5. *Compilation Phase.*

This phase is governed by RECCM, and six other subprograms besides I/O and conversion are called: RECCO, RECFC, RECCA, RECKO, RECCC and RECQU. Each character (of a maximum of 64) is assigned a complication code through ICPL. In order to have more symbols for defining external subroutines and more internal operations the symbol quote is given special treatment. The combination 'x causes the addition of 64 to the RECNE code for symbol x. In this manner access is gained to the upper half of ICPL, declared as a 128 word array.

REC programs are compiled into a special code for the benefit of the interpreter. The simplest signals consist in the sign of the. code words, their being zero, or finally their size. Thus:

    Negative numbers indicate subroutine jumps (consult IXEQ).
    Positive numbers indicate transfers.
    Zero indicates entry points or false exits.
    Very positive numbers (= 2000) indicate recursive return.

There are 14 compilation codes according to the typical code configurations which different characters produce.
These are assigned as follows:

| | |
|---|---|
| left parenthesis | 1 |
| right parenthesis | 2 |
| comma or semicolon | 3 |
| period or colon | 4 |
| operator of no arguments | 5 |
| operator of one numerical argument | 6 |
| predicate of no arguments | 7 |
| predicate with one E13CDIC argument | 8 |
| counter | 9 |
| program constant | 10 |
| quote | 11 |
| program comments | 12 |
| quoted strings | 13 |
| operations found only on CDC 3150 REC | 14 |

    i) RECCM.

On entrance, RECCM tests if IPROG is almost full, and if so, it issues an error message. Otherwise, it reads and echoes a character and gets its six bit code. The compilation code plus one is then put in IX (undefined symbols have zero compilation code) and a transfer is made through a computed GO TO.

RECCM handles comments, quotes and operators defined only in the CDC 3150 version of FORTRAN coded REC (which includes matrix operations). Comment processing consists of reading and writing characters until a quote is found. A quote causes reading and writing of the following character; the EBCDIC code is then reduced to the range 1-64 by RECNE and 64 is added to form the subscript for ICPL. When an operator defined on 3150 REC is found, an error message is output and compilation terminates abnormally.

The rest of the symbols are compiled by the subroutines mentioned before. If an error is detected the output buffer is dumped and a skip to a new page is issued whenever the output device is the line printer. RECCM returns with a negative quantity in IE. The end of compilation is detected by RECCO, which lets RECCM sense this condition by setting IE to zero, in which case IE is reset to the normal status (positive) and a return is executed.

Since semicolons and predicates are associated with a jump to a place in the program unknown when either is found, and track has to be kept of the addresses to which colons will transfer, a pushdown list of triples (IPDL(10,3)) is used. IPDL(IP,l) contains the address on IPROG to which colons on parenthesis level IP have to transfer and false and true exit chains are built on IPDL(IP,2) and IPDL(IP,3) by predicates and semicolons (an the same parenthesis level), respectively.

    ii) RECCO.

    RECCO compiles the control characters (,),: and ;.

Left parentheses produce no coding in IPROG. IP is incremented and tested for overflow, 10 being the deepest parenthesis level allowed; ILC is placed in IPDL(IP,l), supplying thus the address to which colons in level IP will transfer. IPDL(IP,2) and IPDL(IP,3) are set to zero, indicating the end of true exit and false jump chains.

IP is tested for end of program (IP=1) when a right parenthesis is found. If the parenthesis corresponds to a nested expression, a link is built on the outer (IP-1) false exit chain by putting IPDL(IP-1,2) in IPROG(ILC) and ILC in IPDL(IP-1,2): a pointer in IPDL to IPROG(ILC) and zero or a pointer to the next location in IPROG on the chain. If the parenthesis is terminal, these two instructions are skipped. ILC is incremented and RECFC, is called to fill the chains on the current level. IP is decreased to the next level and tested again for termination; if positive, RECCO returns to RECCM, otherwise a zero is placed in IPROG(ILC-1) (the false exit address for the complete expression) and the starting address of the program (ILC0) is put in IPROG(ILC).

The next two characters are read, written, reduced by RECNE and stored in IX and IP (64 being added to IAC before storing in IP); one more character is read and written, the output buffer is dumped and IAC is compared with an L. If equal, the contents of IPROG are dumped from ILC0 through ILC in 11I7 format; this supplies an object listing. If the L is not present, no listing is produced. ILC is incremented, if IX is 1 (a blank in RECNE code) a main REC expression has been found and the main constant storage pointer is set to the value of the subroutine constant storage pointer, IE is set to zero (flagging end of compilation), IL is set to 1 (eliminating list suppression) and a return to RECCM is executed.

If IX is not 1, its compilation code is tested; if this code is equal to 11 (code for '), IP is stored in IX; if not 11, IX is assigned a compiling code of 7 (predicate with no parameters) and, depending on whether it was defined as a recursive subroutine in RECMO (IXEQ(IX)= 500) or not, IXEQ(IX) is set equal to -(ILC0+500) or to –ILC0, respectively. If a recursive subroutine is defined, its entry point is made to contain 2000. The subroutine program constant storage pointer is set to the value of the main constant storage pointer, IP and IPROG(ILC) are set to zero, indicating the beginning of another program, ILC is incremented and blanks are ignored until a left parenthesis is found, in which case a transfer is made to the left parenthesis compilation section. If a character other than a blank or a left parenthesis is found in the space between the subroutine and the next program, an error message is issued.

A semicolon (commas are accepted due to the lack of semicolons on 026 keypunches) creates a link in the true exit chain (IPROG(ILC)=IPDL(I.P,3), IPDL(IP,3)=ILC) and RECFC is called to fill the false transfer chain after ILC has been incremented. IPDL(IP,2) is set to zero to indicate the beginning of a new false chain on the same parenthesis level.

A colon (period) causes the contents of IPDL(IP,1) to be placed in IPROG(ILC), and again ILC is incremented, RECFC is called and IPDL(IP,2) is set to zero.

iii) RECFC.

RECFC(N) fills the chain on IPROG whose start is indicated by N. This is done by recording the value of N; if this value is zero RECFC returns to the calling program. If not zero, this value indicates a cell in IPROG which itself may contain a zero or another link in the chain. The value of ILC is placed in this location after recording its contents, thus establishing the transfer location for execution. The process is repeated until a zero is found in the chain.

iv) RECCA.

This subroutine compile operators and predicates which have no arguments, or which have one digit or single character arguments. Three parameters (N, L, J) are needed in the calling sequence; N indicates the number of arguments, L is the kind of compilation (1 for predicates, 2 for operators), and J specifies the type of argument, 1 indicating a character in Al format (as read by RECR) and 2 meaning a decimal digit. Since digit arguments are used for subscripting, zero is converted to ten.

On entrance, the negative of IAC (which contains the symbol being compiled) is placed in IPROG; any arguments are compiled directly following this cell and a false exit link is established if the compilation corresponds to a predicate. ILC is incremented accordingly.

v) RECKO.

RECKO compiles counters. A counter (specified in the form $n$, where n is an integer) is compiled as a predicate of two arguments, -n being stored in two consecutive locations in IPROG. One of the cells is incremented during execution, and when it reaches zero, the predicate is *false* and the cell is restored with the value -n stored in the other cell. ($5$ o.,) will execute the operation o 5 times (provided there are no predicates in o which are *false*); (o $5$ .,) will execute it 6 times, because the execution has already been carried out before the counter has been incremented, as is clearly obvious from comparing the sequence ($1$ o .,) with (o $1$ .,).

If at the end of the numerical conversion by RECNC n is zero or negative, an error message is issued and compilation terminates abnormally.

vi) RECCC.

Program constants are compiled into the array CONST by RECCC. The character indicating a program constant is compiled as a one argument operator. The parameter specifies the place in CONST where the constant is stored. RECNC is called for the conversion of a floating point number, and a quote is searched after conversion. If it is not found or if more constants than CONST can hold are defined, error messages are provided.

vii) RECQU.

RECQU compiles a quoted string of the form "xxxx'. The operator indicating the quoted string is compiled as an operator with n+1 arguments, n being the number of characters between the operator and the next quote. The word count, n, is placed as the first argument, and the characters in the string are compiled directly into IPROG the way they are read by RECR. ILC is tested for program length.

viii) Compilation Example
.
To illustrate the concept of exit chains, consider the recursive subroutine
$$(N, 0 \ L'/1', P \ '/1' - 'R \ *,)'R$$
When the N (which is a predicate) is seen, the negative of its REC code is put in IPROG(ILC), ILC is incremented, the contents of IPDL(IP,2) (set to zero when the initial left parenthesis was found) are placed in IPROG(ILC), the value of ILC is stored in IPDL(IP,2) and ILC is incremented. The result is illustrated in fig. 1.

The following comma is found next. After setting a link on IPDL(IP,3) and incrementing ILC, RECFC is called to fill the chain starting in IPDL(IP,2), operation whose results are shown in fig. 2.

The contents of IPROG and IPDL after the last comma in the expression is found are illustrated in fig. 3. The string of arrows starting at IPDL(IP,3) shows the true exit chain for the expression. The steps followed by RECFC (which is called when the right parenthesis is found) while filling the chain starting at the address in IPROG stored in IPDL(IP,3) are shown in figures 4(a) through 4(c). The final appearance of IPROG is depicted in fig. 4(d).

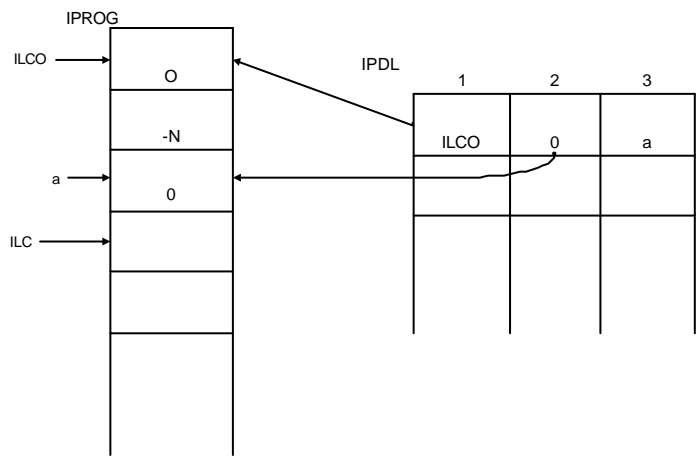

Fig. 1.

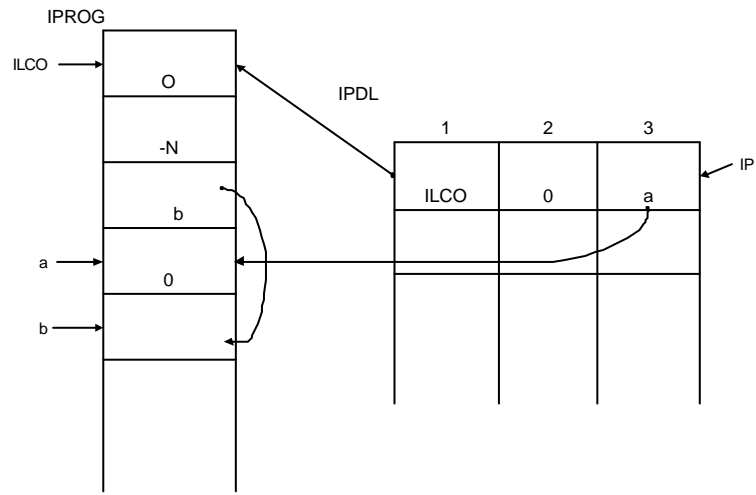

Fig. 2.

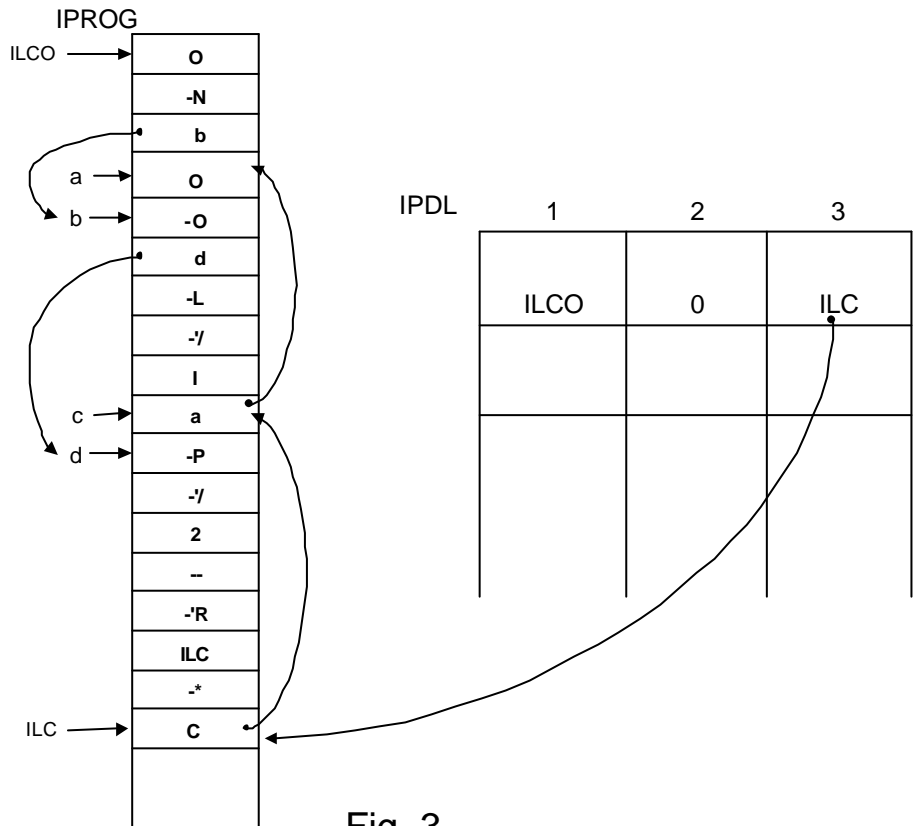

Fig. 3.

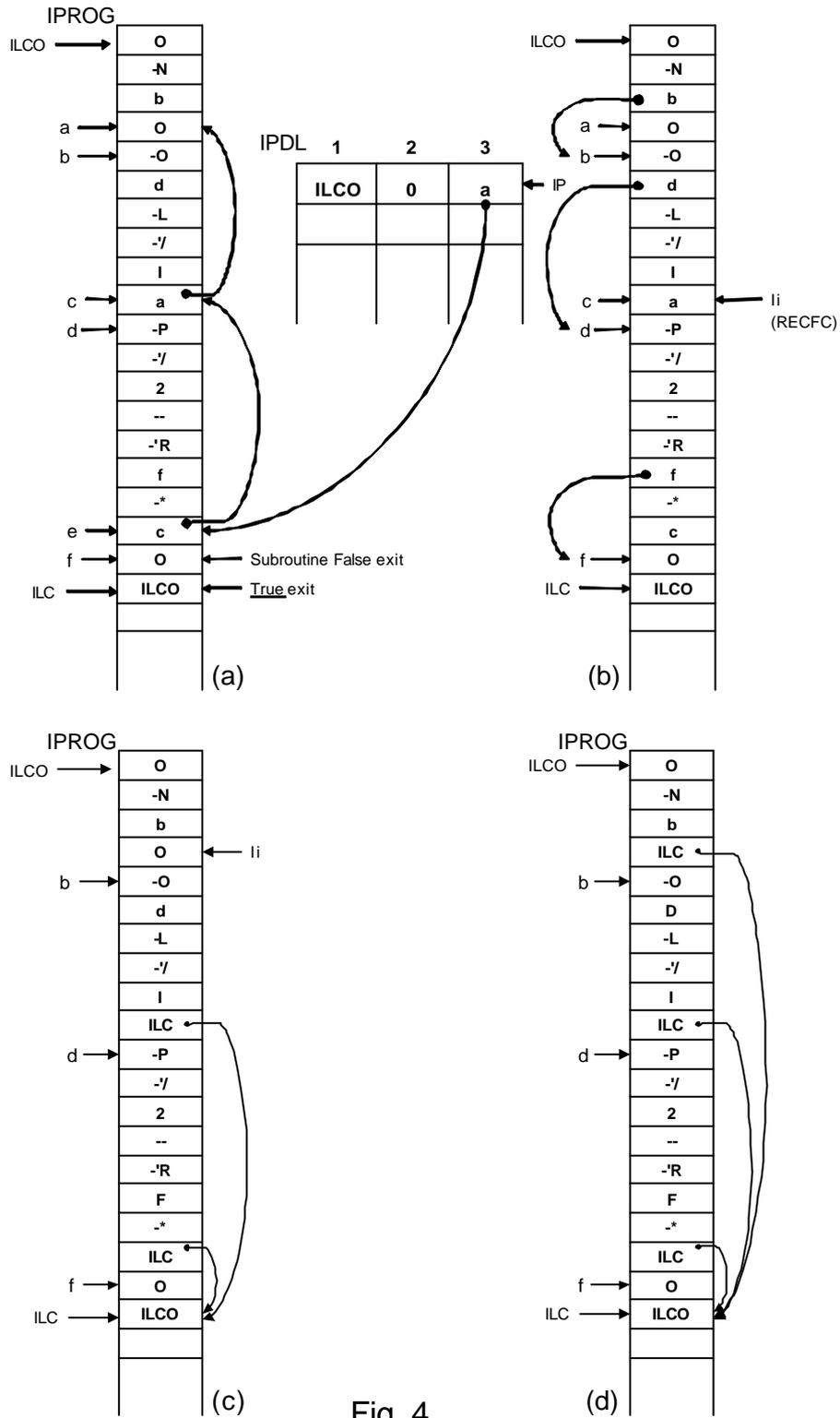

Fig. 4.

6. *Execution Phase.*

The controller for this phase is RECXC. On entrance from the main program,. IXL is set to the value ILC0+1. ILC0 points to the subroutine-type entry address of the last program compiled in the previous phase, which is the main REC expression. The recursion level indicator, IREC, is initialized and the main loop (starting at statement number 1) is entered.

At statement 1, IXL and ILC0 are compared. If they are equal, the entry point of the main program has been reached and its execution is complete; otherwise, IPROG(IXL) is tested. Four kinds of data are distinguished in IPROG:

| | |
|---|---|
| Negative | Subroutine call. |
| Zero | *False* return from a REC defined subroutine. |
| Positive | Unconditional jump. |
| Very positive | (=2000) Entry point of a recursive subroutine. |

When IPROG(IXL) is negative, the execution code of the character represented is obtained from IXEQ(IX), IX being defined previously as -IPROG(IXL). IX is set to the value of the execution code and tested. Five possibilities are found for value of IX:

| | |
|---|---|
| Less than -500 | Recursive subroutine call. |
| Equal to  500 | Recursive subroutine not defined by a REC expression. |
| Negative | REC defined nonrecursive subroutine. |
| Zero | Undefined operation. |
| Positive | System subroutine. |

System subroutines are contained in RECXQ, which uses IX as the index of a computed GO TO. If RECXQ is called, IE is tested after returning. If IE is negative, the program terminates abnormally. The cause of the error is stated by RECXQ before returning.

Entrance to nonrecusive subroutines is accomplished by obtaining the entry point (which is just -IX), storing the true return address at this place and setting IXL to the entry point plus one; control goes back to statement 1.

The entry point address of recursive subroutines is obtained by subtracting 500 from -IX (c. f. RECCO); the recursion pushdown list pointer is tested. If the list is full, a diagnostic is put out and RECXC returns to REC; otherwise the true return address is stored in the list, the pointer is pushed and execution continues at the entry point plus one. Control goes to statement 1 again.

If a zero cell in IPROG is reached, it indicates that the subprogram to which it belongs has terminated with the value false. To calculate the return address, the entry point address is obtained front the word following the zero. The value of the entry point is compared with ILC0 and RECXC returns if they are equal. If ILC0 and the entry point are not equal, the contents of the entry point are tested. If the entry point contains 2000, the return address is fetched from the recursion pushdown list and the pointer is lifted; the false return is obtained by subtracting 1 from the return address. Control goes then to statement 1. A value at the entry point less than 2000 is the actual true return address, so 1 is subtracted to obtain the false return, which is stored in IXL. Once more, control goes to statement 1.

If IPROG(IXL) is positive, IXL is set to this value and tested. If it is equal to 2000, it indicates a true return from a recursive subroutine, so the return address is read from IRET (the recursion pushdown list) and IREC is lifted. If IXL is less than 2000, its value is an address in IPROG (either a true return from a REC defined subroutine or a transfer due to a colon, semicolon or predicate) in which case execution continues at statement 1.

Except for the counter, RECXQ contains all the definitions of the predicates and operators described in section 3. The pushdown list from which tile arithmetic operators get their arguments is operated with a pointer. All the definitions in RECXQ are made in terms of FORTRAN functions and the I/O and conversion subroutines described in section 2. IXL is varied according to the number of arguments of each predicate and error diagnostics are produced in the event of the pushdown list being empty (not enough operands for the process requested), the list being full or data being incorrectly specified. RECCN is in charge of handling counters.

A much shorter version of RECCN can be written in assembly language, using the long MDX instruction. All that RECCN consists of is testing the second word in the counter calling sequence, incrementing it if it is not zero arid returning with the value true, or if the second word is zero, restoring it with the contents of the first words of the calling sequence and returning to the false transfer.

The arithmetic pushdown list can hold up to 500 numbers and a maximum depth of recursion of 100 is provided.

*7. Compiler table Setup[*].*

The tables used by REC are placed on the disk by RECDO and RECER.

RECDO is in charge of reading ICPL, IXEQ and IMON from cards, storing them on the disk and printing them on the console typewriter (fig. 5). The typewriter is used because by suitable changes in the WRTYZ, TYPEZ, HOLEZ, HOLTB and EBCTB library subroutine of the 1130 Disk Monitor System (Version 1), the 64 characters present on the keyboard can be made available. The first card read by RECDO contains these characters in the order they appear when truncated to six bits and are used as a reference on the printed output. The next twelve cards contain the actual tables, four for each table, placed in the order ICPL, IXEQ, IMON. A permanent disk file is allocated with the name RECAT and is equated to the file defined within the program through an *FILES control card.

The error messages used in REC are written by RECER on the permanent disk file RECEM. Twenty cards containing the diagnostic messages are read, written on the disk, read back and put out on the printer (fig. 6).

---

[*] see next note.

```
   A B C D E F G H I ¢ . < ( + | & J K L M N O P Q R ! $ * ) ; ¬
0  5 5 514 5 614 5 5 7 4 2 1 5 7 5 7 7 5 5 5 7 5 5 5 9 9 5 2 3 7
0  114 2 0 321 0 422 0 0 0 015 01518 030 5 6 723 824292917 0 0 0
0  0 0 0 0 4 0 0 0 1 0 0 0 0 0 0 0 0 0 0 0 5 2 0 0 0 0 0 0 0 0

   - / S T U V W X Y Z   , % - > ? 0 1 2 3 4 5 6 7 8 9 : # @ ' = "
   5 5 6141414 5 5 714 7 3 1 7 7 7 7 7 7 7 7 7 7 7 4 81111 813
1619 9 0 0 02528 0 0 0 0 0 0 0 013 0 0 0 0 0 0 0 0 0 027 0 02726
0  0 6 3 0 0 0 0 0 0 0 0 0 0 0 0 0 0 0 0 0 0 0 0 0 0 0 0 0 0 0

   A B C D E F G H I ¢ . < ( + | & J K L M N O P Q R ! $ * ) ; ¬
7  5 7 714 7 7 7 7 7 7 7 7 7 7 714 5 7 7 7 7 7 7 7 712 7 7 7
010 0 0 0 0 0 0 0 0 0 0 0 0 0 0 011 0 0 0 0 0 0 0 0 0 0 0 0 0 0
0  0 0 0 0 0 0 0 0 0 0 0 0 0 0 0 0 0 0 0 0 0 0 0 0 0 0 0 0 0 0

   - / S T U V W X Y Z   , % - > ? 0 1 2 3 4 5 6 7 8 9 : # @ ' = "
710  51414 7 7 7 7 7 7 7 7 7 7 7 7 7 7 7 7 7 7 7 7 7 71313 7 7
02012 0 0 0 0 0 0 0 0 0 0 0 0 0 0 0 0 0 0 0 0 0 0 0 0 02626 0 0
0  0 0 0 0 0 0 0 0 0 0 0 0 0 0 0 0 0 0 0 0 0 0 0 0 0 0 0 0 0 0
```

Fig 5.

```
// JOB T
// XEQ RECER    1
*FILES(2,RECEM)
COMP 01 EXCESS NESTING
COMP 02 PROGRAM LENGTH EXCEEDS CAPACITY
EXEC 01 EXCESSIVE RECURSION
EXEC 02 EMPTY PUSHDOWN LIST
EXEC 03 PUSHDOWN LIST OVERFLOW
COMP 03 ILLEGAL ARGUMENT
COMP 04 ILLEGAL CHARACTER ON PARENTHESIS LEVEL ZERO
COMP 05 NEGATIVE OR ZERO COUNTER
SUP  01 ILLEGAL I/O UNIT NUMBER
COMP 06 PROGRAM DEFINED CONSTANT EXCESS
CONV 01 SYNTAX ERROR IN NUMERIC DATA
EXEC 04 RECURSIVE SUBROUTINE NOT DEFINED
EXEC 05 UNDEFINED NONRECURSIVE SUBROUTINE
REC  01 UNUSED
COMP 07 REC/3150 OPERATOR
REC  02 UNUSED
REC  03 UNUSED
REG  04 UNUSED
REC  05 UNUSED
REC  06 UNUSED
```

Fig 6.

7. *Operation of the system*[*].

  In order to make REC available, the user must define with the DUP *STORE-DATA control card the files mentioned in the last section, allocating two disk sectors to each file.
  It is also convenient to have RECDO and RECER stored in the disk. After this has been done, they are called to execution with the following cards (data cards are included)

```
// XEQ RECDO     1
*FILES(1,RECAT)
 ABCDEFGHI¢.<(+|&JKLMNOPQR!$*);¬-/STUVWXYZ ,%->?0123456789:#@'="
 0 5 5 514 5 614 5 5 7 4 2 1 5 7 5 7 7 5 5 5 7 5 5 5 5 9 9 5 2 3 7
 5 5 6141414 5 5 714 7 3 1 7 7 7 7 7 7 7 7 7 7 7 7 4 81111 813
 7 5 7 714 7 7 7 7 7 7 7 7 7 7 714 7 5 7 7 7 7 7 7 712 7 7 7
 710 51414 7 7 7 7 7 7 7 7 7 7 7 7 7 7 7 7 7 7 7 7 7 71313 7 7
 0 114 2 0 321 0 422 0 0 0 015 01518 030 5 6 723 824292917 0 0 0
1619 9 0 0 02528 0 0 0 0 0 0 013 0 0 0 0 0 0 0 0 027 0 02726
 010 0 0 0 0 0 0 0 0 0 0 0 0 0 0 0 011 0 0 0 0 0 0 0 0 0 0 0
 02012 0 0 0 0 0 0 0 0 0 0 0 0 0 0 0 0 0 0 0 0 0 02626 0 0
 0 0 0 0 0 4 0 0 0 1 0 0 0 0 0 0 0 0 0 0 5 2 0 0 0 0 0 0 0 0
 0 0 6 3 0 0 0 0 0 0 0 0 0 0 0 0 0 0 0 0 0 0 0 0 0 0 0 0 0 0
 0 0 0 0 0 0 0 0 0 0 0 0 0 0 0 0 0 0 0 0 0 0 0 0 0 0 0 0 0 0
 0 0 0 0 0 0 0 0 0 0 0 0 0 0 0 0 0 0 0 0 0 0 0 0 0 0 0 0 0 0

// XEQ RECER     1
*FILES(2,RECEM)
COMP 01 EXCESS NESTING
COMP 02 PROGRAM LENGTH EXCEEDS CAPACITY
EXEC 01 EXCESSIVE RECURSION
EXEC 02 EMPTY PUSHDOWN LIST
EXEC 03 PUSHDOWN LIST OVERFLOW
COMP 03 ILLEGAL ARGUMENT
COMP 04 ILLEGAL CHARACTER ON PARENTHESIS LEVEL ZERO
COMP 05 NEGATIVE OR ZERO COUNTER
SUP  01 ILLEGAL I/O UNIT NUMBER
COMP 06 PROGRAM DEFINED CONSTANT EXCESS
CONV 01 SYNTAX ERROR IN NUMERIC DATA
EXEC 04 RECURSIVE SUBROUTINE NOT DEFINED
EXEC 05 UNDEFINED NONRECURSIVE SUBROUTINE
REC  01 UNUSED
COMP 07 REC/3150 OPERATOR
REC  02 UNUSED
REC  03 UNUSED
REG  04 UNUSED
REC  05 UNUSED
REC  06 UNUSED
```

---

[*] All code FORTRAN REC was compiled on SIMH IBM 1130 Emulator and Disk Monitor System R2V12, and the examples are too .

The REC program and subroutines are next compiled and stored. In a system with 8K of core memory the processor is brought to execution by use of the following control cards:

```
// XEQ REC      2
*LOCALREC,RECCA,RECCC,RECCO,RECKO,RECMO,RECNP,RECQU
*FILES(1,RECAT),(2,RECEM)
```

REC monitor cards, REC programs and data are placed after these cards; the last card in the deck should contain an asterisk in the first column and a T after it.

The *LOCAL card directs the 1130 Monitor System to load the named subroutines only when they are called, reducing thus the core requirements for the processor.

```
// FOR
*NAME REC
*IOCS(CARD,1132 PRINTER,TYPEWRITER,KEYBOARD,DISK)
*ONE WORD INTEGERS
*LIST SOURCE PROGRAM
C *** ***************************************************** ***
C *** REC, THE MAIN PROGRAM. .L READS CHARACTER DEFINIT-    ***
C *** IONS FOR COMPILATION AND EXECUTION FROM DISK          ***
C *** STORAGE IT INITIALIZES 1/0 VARIABLES AND OTHER        ***
C *** POINTERS AND THEN CALLS RECMO AND RECCM FOR           ***
C *** MONITORING AND COMPILATION, RESPECTIVELY. ON          ***
C *** SUCCESSFUL COMPILATION RECXC, THE EXECUTION           ***
C *** INTERPRETER, IS CALLED.                               ***
C *** ***************************************************** ***
      COMMON IAC, ILC, IXL, ILC0
      COMMON IR, IW, IX, IP, IE, IIN, IOUT, IL
      COMMON IPROG(500), IPDL(10, 3), ICPL(128),IXEQ(128)
      COMMON IM,KS,KS1,CONST(30)
      DEFINE FILE 1(3,128, U, KF1), 2(20,30, U, KF2)
      READ(1'1) ICPL
      READ(1'2) IXEQ
      IIN = 2
      IOUT = 3
      IW = 0
      ILC0 = 1
      KS = 0
      KS1 = 0
    1 IM = 1
    2 ILC = ILC0
      IL = 1
      CALL RECMO
      CALL RECCM
```

```
      IF(IE)2,2,3
    3 CALL RECXC
      GO TO 1
C ***                                                        ***
C *** LIST OF AL FORMAT EQUIVALENTS USED IN THE SYSTEM.      ***
C *** SYMBOL     EQUIVALENT  USED IN                         ***
C *** BLANK         16448    RECCC, RECCO, RECNC,            ***
C ***                           RECNP, RECXQ                 ***
C ***   '           32064    RECCC, RECCM, RECQU,            ***
C ***                           RECR, RECXQ                  ***
C ***   L          -11456    RECCO                           ***
C ***   (           19776    RECCO, RECMO, RECR              ***
C ***   *           23616    RECMO                           ***
C ***   C          -15552    RECMO                           ***
C ***   +           20032    RECNC                           ***
C ***   +           20544    RECNC                           ***
C ***   -           24640    RECNC, RECNP                    ***
C ***   .           19264    RECNC, RECNP                    ***
C ***   E          -15040    RECNC, RECNP                    ***
C ***   O           -4032    RECNC, RECNP                    ***
C ***  PERCENT      27712    RECR                            ***
C ***  LOZENGE      19520    RECR                            ***
C ***   )           23872    RECR                            ***
C ***  AT SIGN      31808    RECR                            ***
C ***  NUMBER       31552    RECR                            ***
C ***   =           32320    RECR                            ***
C ***   /           24896    RECXQ                           ***
C ***                                                        ***
      END
FEATURES SUPPORTED
 ONE WORD INTEGERS
 IOCS
CORE REQUIREMENTS FOR REC
 COMMON     862  VARIABLES      16  PROGRAM     90

END OF COMPILATION

// DUP
*DELETE             REC
CART ID 1234   DB ADDR   21B0   DB CNT    0008
*STORE      WS  UA  REC
CART ID 1234   DB ADDR   22B7   DB CNT    0008
```

```
      // FOR
      *ONE WORD INTEGERS
      *LIST SOURCE PROGRAM
            SUBROUTINE RECCA(N,L,J)
      C *** ******************************************************* ***
      C *** RECCA (N, L, J) COMPILES ARGUMENTS FOR PREDICATES        ***
      C *** AND OPERATORS                                            ***
      C *** N = NUMBER OF ARGUMENTS                                  ***
      C *** L = 1 FOR PREDICATES,  = 2   FOR OPERATORS               ***
      C *** J = 1 TYPE  OF ARGUMEN - 1 = EBCDIC, 2 = NUMERICAL       ***
      C *** ******************************************************* ***
            COMMON IAC, ILC, IXL, ILC0
            COMMON IR, IW, IX, IP, IE, IIN, IOUT, IL
            COMMON IPROG(500), IPDL(10, 3), ICPL(128),IXEQ(128)
            COMMON IM,KS,KS1,CONST(30)
      C ***                                                          ***
      C *** IAC CONTAINS THE RECNE CONVERTED CODE OF THE             ***
      C *** CHARACTER REPRESENTING THE SUBROUTINE.                   ***
            IPROG(ILC) = - IAC
            ILC = ILC + 1
            IF(N)30,30,10
      C ***                                                          ***
      C *** INSERT THE ARGUMENTS, CHECKING THEIR VALIDITY            ***
      C *** IF NUMERICAL.                                            ***
         10 DO 20 I = 1,N
            CALL RECR
            CALL RECW(1)
            IPROG(ILC) = IAC
            GO TO (20,60),J
         60 CALL RECNE
      C *** 49 AND 58 ARE THE RECNE CODES FOR 0 AND 9,               ***
      C *** RESPECTIVELY.                                            ***
            IF(IAC-49)65,64,66
         66 IF(IAC-58)67,67,65
         64 IAC = IAC + 10
         67 IPROG(ILC) = IAC-49
         20 ILC = ILC + 1
      C ***                                                          ***
      C *** LINK THE FALSE EXIT IF PREDICATE.                        ***
         30 GO TO (40,50),L
         40 IPROG(ILC) = IPDL(IP, 2)
            IPDL (IP,2) = ILC
            ILC = ILC + 1
         50 RETURN
      C ***                                                          ***
      C *** ERROR CONDITIONS                                         ***
```

```
   65 IE=-6
      CALL RECW(3)
      RETURN
      END
FEATURES SUPPORTED
 ONE WORD INTEGERS
CORE REQUIREMENTS FOR RECCA
 COMMON      862 VARIABLES      4 PROGRAM     156
RELATIVE ENTRY POINT ADDRESS IS 000A (HEX)
END OF COMPILATION
// DUP
*DELETE             RECCA
CART ID 1234   DB ADDR   21B0   DB CNT    000A
*STORE       WS  UA  RECCA
CART ID 1234   DB ADDR   22B5   DB CNT    000A
// FOR
*ONE WORD INTEGERS
*LIST SOURCE PROGRAM
      SUBROUTINE RECCC
C *** ***************************************************** ***
C *** RECCC COMPILES PROGRAM CONSTANTS INTO CONST            ***
C *** ***************************************************** ***
      COMMON IAC,ILC,IXL,ILC0
      COMMON IR,IW,IX,IP,IE,IIN,IOUT,IL
      COMMON IPROG(500),IPDL(10,3),ICPL(128),IXEQ(128)
      COMMON IM,KS,KS1,CONST(30)
C ***                                                        ***
C *** ENTER CODE FOR THE ROUTINE WHICH PLACES A              ***
C *** PROGRAM DEFINED CONSTANT ON THE PUSHDOWN LIST          ***
C *** (REC SYMBOL'/).                                        ***
      IPROG(ILC) = -IAC
      ILC = ILC + 1
C ***                                                        ***
C *** READ FLOATING POINT CONSTANT TO G --                   ***
C ***    '/ DEFINES THE CONSTANT,                            ***
C ***    ANY NUMBER OF BLANKS MAY PRECEDE THE                ***
C ***                                         NUMBER,        ***
C ***    SPACES FOLLOWING THE NUMBER ARE ALSO                ***
C ***                                         IGNORED,       ***
C ***    ' TERMINALES INPUT OF THE NUMBER, ANY OTHER         ***
C ***                                         SYMBOL         ***
C ***    CAUSES AN ERROR                                     ***
      CALL RECNC(1,G,I)
    1 IF(IAC-16448)9,3,9
    3 CALL RECR
      CALL RECW(1)
      GO TO 1
    9 IF(IAC-32064)4,5,4
C ***                                                        ***
```

```
C *** RECORD THE CONSTANT IN THE CONSTANT ARRAY.                ***
C *** RECORD THE LOCATION OF THE PROGRAM CONSTANT               ***
C *** IN THE CALLING SEQUENCE OF '/.                            ***
C *** OUTPUT ERROR MESSAGE IF ARRAY OVERFLOWED.                 ***
    5 KS=KS+1
      IPROG(ILC)=KS
      ILC=ILC+1
      IF(KS-30)6,6,7
    6 CONST(KS)=G
      RETURN
C ***                                                           ***
C *** ERROR CONDITIONS                                          ***
    4 IE=-11
      GO TO 8
    7 IE=-10
    8 CALL RECW(3)
      RETURN
      END
FEATURES SUPPORTED
 ONE WORD INTEGERS
CORE REQUIREMENTS FOR RECCC
 COMMON     862 VARIABLES       4 PROGRAM     104
RELATIVE ENTRY POINT ADDRESS IS 000B (HEX)
END OF COMPILATION
// DUP
*DELETE              RECCC
CART ID 1234   DB ADDR   21C0    DB CNT    0007
*STORE        WS  UA  RECCC
CART ID 1234   DB ADDR   22C9    DB CNT    0007
// FOR
*ONE WORD INTEGERS
*LIST SOURCE PROGRAM
      SUBROUTINE RECCM
C *** ****************************************************** ***
C *** RECCM GETS COMPILATION CODES AND TRANSFERS TO          ***
C *** THE APPROPRIATE COMPILING SUBPOUTINE                   ***
C *** ****************************************************** ***
      COMMON IAC,ILC,IXL,ILC0
      COMMON IR,IW,IX,IP,IE,IIN,IOUT,IL
      COMMON IPROG(500),IPDL(10,3),ICPL(128),IXEQ(128)
      COMMON IM,KS,KS1,CONST(30)
   10 IF(ILC-495)4,4,5
    4 CALL RECR
      CALL RECW(1)
      CALL RECNE
    3 IX=ICPL(IAC)+1
C ***              ( ) ,  . O0 O1 P0 PH  $ '/  '  '*  ''  CDC ***
      GO TO(10,20,20,20,20,60,61,70,73,80,90,100,101,102,110),IX
C ***                                                           ***
```

```
C *** CONTROL CHARACTERS ( ) , .                                 ***
   20 CALL RECCO
  150 IF(IE)81,190,10
  190 IE = 1
  191 RETURN
   81 CALL RECW(2)
      IF(IOUT-3)191,82,191
   82 WRITE(3,300)
      GO TO 191
  300 FORMAT('1')
C ***                                                            ***
C *** OPERATORS WITH NO ARGUMENTS                                ***
   60 CALL RECCA(0,2,2)
      GO TO 150
C ***                                                            ***
C *** OPERATORS WITH ONE NUMERICAL ARGUMENT                      ***
   61 CALL RECCA(1,2,2)
      GO TO 150
C ***                                                            ***
C *** PREDICATES WITH NO ARGUMENTS                               ***
   70 CALL RECCA(0,1,2)
      GO TO 150
C ***                                                            ***
C *** PREDICATES WITH ONE HOLLERITH ARGUMENT                     ***
   73 CALL RECCA(1,1,1)
      GO TO 10
C ***                                                            ***
C *** COUNTERS ($N$)                                             ***
   80 CALL RECKO
      GO TO 150
C ***                                                            ***
C *** PROGRAM CONSTANTS  ('/XXX')                                ***
   90 CALL RECCC
      GO TO 150
C ***                                                            ***
C *** INDEXING FOR QUOTED CHARACTERS - ADD 64 TO USE             ***
C *** UPPER HALF                                                 ***
C *** UPPER HALF OF TABLE                                        ***
  100 CALL RECR
      CALL RECW(1)
      CALL RECNE
      IAC=IAC+64
      GO TO 3
C ***                                                            ***
C *** COMMENTS ('*XXX')                                          ***
  101 CALL RECR
      CALL RECW(1)
      IF(IAC-32064)101,4,101
```

```
C ***                                                     ***
C *** QUOTED STRINGS (''XXX')                             ***
  102 CALL RECQU
      GO TO 150
C ***                                                     ***
C *** OPERATORS DEFINED ONLY ON CDC 3150 REC              ***
C ***                                                     ***
C *** ERROR CONDITIONS                                    ***
  110 IE=-15
      GO TO 194
    5 IE=-2
  194 CALL RECW(3)
      GO TO 81
      END
FEATURES SUPPORTED
 ONE WORD INTEGERS
CORE REQUIREMENTS FOR RECCM
 COMMON     862 VARIABLES       2 PROGRAM     170
RELATIVE ENTRY POINT ADDRESS IS 000D (HEX)
END OF COMPILATION
// DUP
*DELETE            RECCM
CART ID 1234   DB ADDR  21C0   DB CNT    000B
*STORE      WS  UA  RECCM
CART ID 1234   DB ADDR  22C5   DB CNT    000B
// FOR
*ONE WORD INTEGERS
*LIST SOURCE PROGRAM
      SUBROUTINE RECCN
C *** ***************************************************** ***
C *** RECCN EXECUTES COUNTERS                              ***
C *** ***************************************************** ***
      COMMON IAC,ILC,IXL,ILC0
      COMMON IR,IW,IX,IP,IE,IIN,IOUT,IL
      COMMON IPROG(500),IPDL(10,3),ICPL(128),IXEQ(128)
      COMMON IM,KS,KS1,CONST(30)
C ***                                                     ***
C *** REC CODE COMPILED FROM THE COUNTER $N$ HAS THE      ***
C *** FORM                 -$     (-28)                   ***
C ***                      -N                             ***
C ***                       K                             ***
C ***                        FALSE                        ***
C ***                         TRUE.                       ***
C *** K=-N INITIALLY. ON EACH CALL TO $ IT IS TESTED      ***
C *** AND THEN INCREMENTED. IF K IS NOT ZERO WHEN         ***
C *** TESTED, THE TRUE EXIT IS TAKEN, IF K IS ZERO IT IS  ***
```

```
C *** RESET TO -N AND THE FALSE EXIT TAKEN                      ***
      IXL=IXL+1
      IF(IPROG(IXL))2,3,3
    3 IPROG(IXL)=IPROG(IXL-1)
      IXL=IXL+1
      RETURN
    2 IPROG(IXL)=IPROG(IXL)+1
      IXL=IXL+2
      RETURN
      END
FEATURES SUPPORTED
 ONE WORD INTEGERS
CORE REQUIREMENTS FOR RECCN
 COMMON     862 VARIABLES        2 PROGRAM      62
RELATIVE ENTRY POINT ADDRESS IS 0004 (HEX)
END OF COMPILATION
// DUP
*DELETE            RECCN
CART ID 1234   DB ADDR  21C0   DB CNT   0005
*STORE       WS  UA  RECCN
CART ID 1234   DB ADDR  22CB   DB CNT   0005
// FOR
*ONE WORD INTEGERS
*LIST SOURCE PROGRAM
      SUBROUTINE RECCO
C *** ***************************************************** ***
C *** RECCO COMPILES CONTROL CHARACTERS AND DOES             ***
C *** FINAL PROCESSING WHEN THE LAST RIGHT PARENTHE-         ***
C *** SIS OF A PROGRAM IS FOUND                              ***
C *** ***************************************************** ***
      COMMON IAC,ILC,IXL,ILC0
      COMMON IR,IW,IX,IP,IE,IIN,IOUT,IL
      COMMON IPROG(500),IPDL(10,3),ICPL(128),IXEQ(128)
      COMMON IM,KS,KS1,CONST(30)
  300 FORMAT (11I7)
      IX=IX-1
C ***        ( ) , .                                         ***
      GO TO (1,2,3,4),IX
C ***                                                        ***
C ***                                                        ***
C *** LEFT PARENTHESES                                       ***
C ***    ADVANCE PARENTHESIS LEVEL                           ***
C ***    TEST MAXIMUM DEPTH                                  ***
C ***    RECORD COLON'S TRANSFER ADDRESS                     ***
C ***    SET UP NULL EXIT CHAINS                             ***
    1 IP=IP+1
      IF(IP-10)22,22,21
   22 IPDL(IP,1)=ILC
      IPDL(IP,2)=0
```

```
      IPDL(IP,3)=0
      RETURN
C ***                                                     ***
C *** RIGHT PARENTHESES                                   ***
C ***   LINK FALSE EXIT ON UPPER LEVEL IF NOT LEVEL       ***
C ***   ZERO                                              ***
C ***   FILL WAITING EXIT CHAINS                          ***
C ***   LIFT LEVEL                                        ***
C ***   TEST FOR FINAL PARENTHESIS                        ***
    2 IF(IP-1)30,30,31
   31 IPROG(ILC)=IPDL(IP-1,2)
      IPDL(IP-1,2)=ILC
   30 ILC = ILC + 1
      CALL RECFC(IPDL(IP,2))
      CALL RECFC(IPDL(IP,3))
      IP=IP-1
      IF(IP)190,190,10
C ***                                                     ***
C *** CLOSING RIGHT PARENTHESIS PROCESSING                ***
C ***     WRITE TERMINAL CODE FOR SUBROUTINE EXIT         ***
C ***     READ FOLLOWING 2 CHARACTERS&&                   ***
C ***       BLANK         IMMEDIATE EXECUTION             ***
C ***       X OR 'X       DEFINE SUBROUTINE               ***
  190 IPROG(ILC-1)=0
      IPROG(ILC)=ILC0
      CALL RECR
      CALL RECW(1)
      CALL RECNE
      IX=IAC
      CALL RECR
      CALL RECW(1)
      CALL RECNE
      IP=IAC+64
C ***         PRODUCE OBJECT LIST IF L IS PRESENT         ***
      CALL RECR
      CALL RECW(1)
      CALL RECW(2)
      IF(IAC + 11456)193,194,193
  194 WRITE(IOUT,300)(IPROG(I),I=ILC0,ILC)
  193 ILC=ILC+1
C *** 1 IS THE RECNE CODE FOR BLANK.                      ***
      IF(IX-1) 191,192,191
C ***                                                     ***
C *** DEFINE SUBROUTINE                                   ***
  191 IF(ICPL(IX)-11)196,195,196
  195 IX=IP
  196 ICPL(IX)=7
      IF(IXEQ(IX)+500)199,197,199
C ***          RECURSIVE SUBROUTINE                       ***
```

```
  197 IXEQ(IX)=-ILC0-500
      IPROG(ILC0)=2000
      GO TO 198
C ***          NONRECURSIVE SUBROUTINE                           ***
  199 IXEQ(IX)=-ILC0
C ***                                                            ***
C *** PREPARE TO COMPILE SUBSEQUENT REC EXPRESSION.              ***
C *** COMPILING CYCLE CONTINUES UNTIL AN EXECUTABLE              ***
C *** EXRESSION IS ENCOUNTERED                                   ***
  198 KS1=KS
      IP=0
      ILC0=ILC
      IPROG(ILC)=0
      ILC=ILC+1
   35 CALL RECR
      IF(IAC-19776)33,36,33
   33 IF(IAC-16448)34,35,34
   36 CALL RECW(1)
      GO TO 1
C ***                                                            ***
C *** IMMEDIATE EXECUTION                                        ***
  192 KS=KS1
      IE=0
      IL=1
   10 RETURN
C ***                                                            ***
C *** COMMAS (SEMICOLONS)                                        ***
C ***    LINK INTO TRUE EXIT CHAIN                               ***
C ***    FILL PREDICATE CHAIN (GO TO 5)                          ***
    3 IPROG(ILC)=IPDL(IP,3)
      IPDL(IP,3) = ILC
      GO TO 5
C ***                                                            ***
C *** PERIODS (COLONS)                                           ***
C    INSERT RETURN TRANSFER                                      ***
C    FILL WAITING PREDICATE CHAIN                                ***
    4 IPROG(ILC)=IPDL(IP,1)
    5 ILC=ILC+1
      CALL RECFC(IPDL(IP,2))
      IPDL(IP,2)=0
      RETURN
C ***                                                            ***
C *** ERROR CONDITIONS                                           ***
   21 IE=-1
      GO TO 32
   34 IE=-7
   32 CALL RECW(3)
      RETURN
      END
```

```
FEATURES SUPPORTED
 ONE WORD INTEGERS
CORE REQUIREMENTS FOR RECCO
 COMMON      862  VARIABLES       4  PROGRAM      446
RELATIVE ENTRY POINT ADDRESS IS 0014 (HEX)
END OF COMPILATION

// DUP
*DELETE            RECCO
CART ID 1234   DB ADDR  21C0   DB CNT   001C
*STORE       WS  UA  RECCO
CART ID 1234   DB ADDR  22B4   DB CNT   001C
// FOR
*ONE WORD INTEGERS
*LIST SOURCE PROGRAM
      SUBROUTINE RECDS(K,L)
C *** ***************************************************** ***
C *** RECDS TESTS DATA SWITCH K RECORDING THE RESULT         ***
C *** IN L. IF THE SWITCH IS ON, RECDS OUTPUTS A MESSAGE     ***
C *** ON BOTH THE CONSOLE PRINTER AND THE CURRENT            ***
C *** OUTPUT DEVICE.                                         ***
C *** ***************************************************** ***
      COMMON IAC,ILC,IXL,ILC0
      COMMON IR,IW,IX,IP,IE,IIN,IOUT,IL
      COMMON IPROG(500),IPDL(10,3),ICPL(128),IXEQ(128)
      COMMON IM,KS,KS1,CONST(30)
  300 FORMAT(' MANUAL INTERRUPT FROM SWITCH ',I2)
      CALL DATSW(K,L)
      GO TO (1,2),L
    1 WRITE(1,300)K
      WRITE (IOUT,300)K
    2 RETURN
      END
FEATURES SUPPORTED
 ONE WORD INTEGERS
CORE REQUIREMENTS FOR RECDS
 COMMON      862  VARIABLES       0  PROGRAM       54
RELATIVE ENTRY POINT ADDRESS IS 0013 (HEX)
END OF COMPILATION

// DUP
*DELETE            RECDS
CART ID 1234   DB ADDR  21C0   DB CNT   0005
*STORE       WS  UA  RECDS
CART ID 1234   DB ADDR  22CB   DB CNT   0005
// FOR
```

```
      *ONE WORD INTEGERS
      *LIST SOURCE PROGRAM
            SUBROUTINE RECFC(N)
C *** ******************************************************   ***
C *** RECFC(N) FILLS THE EXIT CHAIN STARTING AT N WITH          ***
C *** THE CURRENT VALUE OF ILC. THE CHAIN IS TERMINA-           ***
C *** TED BY A CELL CONTAINING A ZERO, IN WHICH ILC IS          ***
C *** ALSO WRITTEN. IF N = 0, RETURN IS IMMEDIATE.              ***
C *** ******************************************************   ***
            COMMON IAC,ILC,IXL,ILC0
            COMMON IR,IW,IX,IP,IE,IIN,IOUT,IL
            COMMON IPROG(500),IPDL(10,3),ICPL(128),IXEQ(128)
            COMMON IM,KS,KS1,CONST(30)
            I1=N
          1 IF(I1)2,3,2
          2 I2=I1
            I1=IPROG(I2)
            IPROG(I2)=ILC
            GO TO 1
          3 RETURN
            END
      FEATURES SUPPORTED
       ONE WORD INTEGERS
      CORE REQUIREMENTS FOR RECFC
       COMMON     862 VARIABLES        4 PROGRAM      36
      RELATIVE ENTRY POINT ADDRESS IS 0004 (HEX)
      END OF COMPILATION
      // DUP
      *DELETE              RECFC
      CART ID 1234   DB ADDR   21C0   DB CNT    0003
      *STORE       WS  UA  RECFC
      CART ID 1234   DB ADDR   22CD   DB CNT    0003
      // FOR
      *ONE WORD INTEGERS
      *LIST SOURCE PROGRAM
            SUBROUTINE RECKO
C *** ******************************************************   ***
C *** RECKO COMPILES COUNTERS                                   ***
C *** ******************************************************   ***
            COMMON IAC,ILC,IXL,ILC0
            COMMON IR,IW,IX,IP,IE,IIN,IOUT,IL
            COMMON IPROG(500),IPDL(10,3),ICPL(128),IXEQ(128)
            COMMON IM,KS,KS1,CONST(30)
C ***                                                           ***
C *** THE CONFIGURATION FOR $N$ WHICH THIS SUB-                 ***
C *** ROUTINE COMPILES IS    -$    (-28)                        ***
C ***                                -N                         ***
C ***                                FALSE.                     ***
            IPROG(ILC)=-IAC
```

```
      ILC=ILC+1
      CALL RECNC(2,G,ICO)
      IF(ICO)82,82,83
   82 IE=-8
      CALL RECW(3)
      RETURN
   83 IPROG(ILC)=-ICO
      IPROG(ILC+1)=-ICO
      ILC=ILC + 2
      IPROG(ILC)=IPDL(IP,2)
      IPDL(IP,2)=ILC
      ILC=ILC+1
      RETURN
      END
FEATURES SUPPORTED
 ONE WORD INTEGERS
CORE REQUIREMENTS FOR RECKO
 COMMON     862 VARIABLES        6 PROGRAM     96
RELATIVE ENTRY POINT ADDRESS IS 000A (HEX)
END OF COMPILATION
// DUP
*DELETE             RECKO
CART ID 1234   DB ADDR   21C0    DB CNT    0007
*STORE      WS UA RECKO
CART ID 1234   DB ADDR   22C9    DB CNT    0007
// FOR
*LIST SOURCE PROGRAM
*ONE WORD INTEGERS
      SUBROUTINE RECMO
C *** *****************************************************  ***
C *** RECMO READS MONITOR CONTROL CHARACTERS AND              ***
C *** INITIALIZES FOR COMPILATION WHEN A LEFT                 ***
C *** PARENTHESIS IS FOUND                                    ***
C *** *****************************************************  ***
      DIMENSION IMON(64)
      COMMON IAC,ILC,IXL,ILC0
      COMMON IR,IW,IX,IP,IE,IIN,IOUT,IL
      COMMON IPROG(500),IPDL(10,3),ICPL(128),IXEQ(128)
      COMMON IM,KS,KS1,CONST(30)
      READ(1'3)IMON
C ***                                                         ***
C *** INITIALIZE                                              ***
C ***    FETCH MONITOR DIRECTORY                              ***
C ***    READ A FRESH CARD                                    ***
C ***    LOOK FOR C OR * IN COLUM 1, IGNORING CARD            ***
C ***    WHEN ABSTENT.                                        ***
      IE=1
   20 IR=81
      CALL RECR
```

```
      CALL RECW(1)
      IF(IAC-23616)19,52,19
   19 IF(IAC+15552)14,50,14
   14 IW=0
      GO TO 20
C ***                                                           ***
C *** A CARD WITH C IN COLUMN 1 CONTAINS COMMENTS. IT           ***
C *** IS WRITTEN IN ITS ENTIRETY.                               ***
   50 DO 51 I=1,79
      CALL RECR
   51 CALL RECW(1)
      CALL RECW(2)
      GO TO 20
C ***                                                           ***
C *** A CARD WITH AN * IN COLUMN 1 INDICATES THAT               ***
C *** MONITOR COMMANDS AND/OR A PROGRAM FOLLOWS                 ***
   52 CALL RECR
C ***                                                           ***
C *** A LEFT PARENTHESIS TERMINATES THE MONITOR                 ***
C *** PHASE.                                                    ***
      IF(IAC-19776)16,15,16
   16 CALL RECW(1)
      CALL RECNE
      IX=IMON(IAC)
C ***                                                           ***
C *** UNDEFINED SYMBOLS ARE PASSED OVER, A VALID ONE            ***
C *** CAUSES THE READING OF THE NEXT CHARACTER FOR              ***
C *** USE AS AN ARGUMENT.                                       ***
      IF(IX)52,52,35
   35 CALL RECR
      IF(IAC-19776)30,15,30
   30 CALL RECW(1)
      CALL RECNE
C ***                                                           ***
C ***        I O T E N S                                        ***
      GO TO (1,2,3,4,5,6),IX
C ***                                                           ***
C ***                                                           ***
C *** IK MAKES THE CURRENT INPUT DEVICE BE UNIT                 ***
C *** NUMBER K, ACCORDING TO 1130 FORTRAN CONVENTIONS           ***
C *** 49,51 AND 55 ARE THE RECNE CODES FOR 0, 2 AND 6,          ***
C *** RESPECTIVELY                                              ***
    1 IF(IAC-55)10,11,12
   10 IF(IAC-51)12,11,12
   11 IIN=IAC-49
      GO TO 52
```

```
   12 IE=-9
      CALL RECW(3)
      IE=1
      GO TO 52
C ***                                                        ***
C *** OK DEFINES THE CURRENT OUTPUT DEVICE TO BE              ***
C *** UNIT NUMBER K. THE RECNE CODES FOR I AND 3 ARE 50       ***
C *** AND 52. RESPECTIVELY.                                   ***
    2 IF(IAC-52)22,23,12
   22 IF(IAC-50)12,23,23
   23 IOUT=IAC-49
      GO TO 52
C ***                                                        ***
C *** T TERMINATES THE RUN BY CALLING EXIT                    ***
    3 CALL RECW(2)
      CALL EXIT
C ***                                                        ***
C *** E ERASES THE CONTENTS OF IPROG BY RETURNING ILC         ***
C *** TO 1 AND RESTORES ICPL AND IXEQ, ERASING ANY DE-        ***
C *** FINITIONS WHICH WERE MADE.                              ***
    4 ILC=1
      READ(1'1)ICPL
      READ(1'2)IXEQ
      KS=0
      KS1=0
      GO TO 52
C ***                                                        ***
C *** NX OR N'X DECLARE X OR 'X RESPECTIVELY AS A             ***
C *** RECURSIVE PREDICATE.                                    ***
    5 IF(ICPL(IAC)-11)53,54,53
   54 CALL RECR
      CALL RECW(1)
      CALL RECNE
      IAC=IAC+64
   53 ICPL(IAC)=7
      IXEQ(IAC)=-500
      GO TO 52
C ***                                                        ***
C *** S SUPPRESSES RECW(1) ACTION DURING COMPILATION.         ***
C *** NO FURTHER OUTPUT WILL BE GENERATED UNTIL THE           ***
C *** EXECUTION OF A PROGRAM, ALTHOUGHTHE WAITING             ***
C *** LINE WILL NOT BE WRITTEN UNTIL A LEFT PARENTHE-         ***
C *** SIS IS FOUND (STATEMENT 15).                            ***
    6 IL=2
      GO TO 52
C ***                                                        ***
C *** INITIALIZATION                                          ***
```

```
C *** SERVE SPACE SINCE RECCM IS NOT PALACED IN LOCAL
   15 IP=1
      CALL RECW(1)
      CALL RECW(2)
      ILC0=ILC
      IPROG(ILC)=0
      ILC=ILC+1
      IPDL(IP,1)=ILC
      IPDL(IP,2)=0
      IPDL(IP,3)=0
      RETURN
      END
FEATURES SUPPORTED
 ONE WORD INTEGERS
CORE REQUIREMENTS FOR RECMO
 COMMON     862 VARIABLES      68 PROGRAM     324
RELATIVE ENTRY POINT ADDRESS IS 0057 (HEX)
END OF COMPILATION

// DUP
*DELETE             RECMO
CART ID 1234   DB ADDR   21C0    DB CNT    0014
*STORE       WS   UA  RECMO
CART ID 1234   DB ADDR   22BC    DB CNT    0014
// FOR
*ONE WORD INTEGERS
*LIST SOURCE PROGRAM
      SUBROUTINE RECNC(JJ,FCONV,ICONV)
C *** ****************************************************** ***
C *** RECNC CONVERTS A STRING OF EBCDIC CHARACTERS           ***
C *** INTO A FIXED OR FLOATING POINT NUMBER.                 ***
C *** JJ=0 INDICATES FLOATING POINT CONVERSION WITH-         ***
C *** OUT ECHOING THE CHARACTERS READ, JJ=1 WILL PRO-        ***
C *** VIDE F. P. CONVERSION WITH WRITING AND JJ=2 WILL       ***
C *** CAUSE THE CONVERSION OF AN INTEGER QUANTITY            ***
C *** WHICH ALWAYS IS DONE WITH CHARACTER ECHOING.           ***
C *** ICONV WILL RETURN THE RESULT OF AN INTEGER CON-        ***
C *** VERSION, FCONV RETURNS THE VALUE OF A FLOATING         ***
C *** POINT CONVERSION.                                      ***
C *** ****************************************************** ***
      COMMON IAC,ILC,IXL,ILC0
      COMMON IR,IW,IX,IP,IE,IIN,IOUT,IL
      COMMON IPROG(500),IPDL(10,3),ICPL(128),IXEQ(128)
      COMMON IM,KS,KS1,CONST(30)
      I=JJ
      FCONV=0.0
      ICONV=0
      IF(I)2,2,10
   10 GO TO (1,4),I
```

```
C ***                                                             ***
C *** FLOATING POINT CONVERSION                                   ***
C ***                                                             ***
    2 I=IL
      IL=2
      GO TO 50
    1 I=IL
   50 JEXP=0
      K=0
      SGN=1
      ISGN=1
    5 CALL RECR
      CALL RECW(1)
C ***                                                             ***
C *** IGNORE LEADING BLANKS                                       ***
      IF(IAC-16448)7,5,7
C ***                                                             ***
C *** TEST FOR MINUS SIGN, PLUS SIGN, AND AMPERSAND               ***
    7 IF(IAC-24640)15,30,15
   15 IF(IAC-20032)52,26,52
   52 IF(IAC-20544)19,26,19
   30 SGN=-1
   26 CALL RECR
      CALL RECW(1)
   19 IF(K)40,40,3
C ***                                                             ***
C *** TEST FOR DECIMAL POINT                                      ***
   40 IF(IAC-19264)9,6,9
    6 K=K+1
      GO TO 26
    3 K=K+1
C ***                                                             ***
C *** TEST FOR E                                                  ***
    9 IF(IAC+15040)8,13,8
C ***                                                             ***
C *** EBCDIC CODES FOR DIGITS LIE IN THE INTERVAL                 ***
C ***  (0,-4032)                                                  ***
    8 IF(IAC)14,12,12
   14 IF(IAC+4032)12,11,11
   11 FCONV=FCONV* 10.+FLOAT((IAC+4032)/256)
      GO TO 26
C ***                                                             ***
C *** EXPONENT CONVERSION                                         ***
   13 CALL RECR
      CALL RECW(1)
C ***                                                             ***
C *** SIGN TEST                                                   ***
      IF(IAC-24640)49,17,49
```

```
   49 IF(IAC-20032)53,35,53
   53 IF(IAC-20544)16,35,16
   17 ISGN=-1
   35 CALL RECR
      CALL RECW(1)
C ***                                                              ***
C *** DIGIT TEST                                                   ***
   16 IF(IAC)18,12,12
   18 IF(IAC+4032)12,20,20
   20 JEXP=10*JEXP+(IAC+4032)/256
      GO TO 35
   12 IF(K)21,21,22
   22 K=K-2
   21 JEXP=ISGN*JEXP-K
      FCONV=SGN*FCONV*10.**JEXP
      IL=I
      RETURN
C ***                                                              ***
C *** INTEGER CONVERSION                                           ***
C ***                                                              ***
    4 ISGN=1
   37 CALL RECR
      CALL RECW(1)
C ***                                                              ***
C *** BLANK TEST                                                   ***
      IF(IAC-16448)36,37,36
C ***                                                              ***
C *** SIGN TEST                                                    ***
   36 IF(IAC-24640)38,27,38
   38 IF(IAC-20032)54,31,54
   54 IF(IAC-20544)25,31,25
   27 ISGN=-1
   31 CALL RECR
      CALL RECW(1)
C ***                                                              ***
C *** DIGIT TEST                                                   ***
   25 IF(IAC)32,33,33
   32 IF(IAC+4032)33,34,34
   34 ICONV=ICONV*10+(IAC+4032)/256
      GO TO 31
   33 ICONV=ISGN*ICONV
      RETURN
      END
FEATURES SUPPORTED
 ONE WORD INTEGERS
CORE REQUIREMENTS FOR RECNC
 COMMON      862 VARIABLES        10 PROGRAM      380
RELATIVE ENTRY POINT ADDRESS IS 001A (HEX)
END OF COMPILATION
```

```
// DUP
*DELETE              RECNC
CART ID 1234   DB ADDR  21C0    DB CNT    0018
*STORE       WS UA  RECNC
CART ID 1234   DB ADDR  22B8    DB CNT    0018
// ASM
*LIST
                       00001      * *** **************************** ***
                       00002      * ***RECNE CONVERTS A LEFT JUSTIFIED***
                       00003      * *** EBCDIC CHARACTER INTO A 6 BIT ***
                       00004      * *** CODE USABLE FOR SUBSCRIPTING.  ***
                       00005      * *** FORTRAN DOES NOT ALLOW ZERO    ***
                       00006      * *** SUBSCRIPTS SO A ONE IS ADDED   ***
                       00007      * *** TO THE RESULT                  ***
                       00008      * *** **************************** ***
0000   19143545        00009             ENT     RECNE
0000 0 0000            00010      RECNE  DC      0
0001 00 C4007FFF       00011             LD    L IAC
0003 0 1002            00012             SLA     2
0004 0 180A            00013             SRA     10
0005 00 D4007FFF       00014             STO   L IAC
0007 00 74017FFF       00015             MDX   L IAC,1
0009 01 4C800000       00016             BSC   I RECNE
                       00017      * ***                                ***
                       00018      * *** ADDRESS OF IAC IN FORTRAN COM-***
                       00019      * *** MON AREA                       ***
7FFF                   00020      IAC    EQU     /7FFF
                       00021      * ***                                ***
                       00022      * ***                                ***
                       00023      * *** A FORTRAN PROGRAM THAT PRODU-  ***
                       00024      * *** CES SIMILAR RESULTS WOULD BE   ***
                       00025      * ***                                ***
                       00026      * ***                                ***
                       00027      * *** // FOR                         ***
                       00028      * *** *ONE WORD INTEGERS             ***
                       00029      * ***       SUBROUTINE RECNE         ***
                       00030      * ***       COMMON IAC               ***
                       00031      * ***       IF(IAC)2,2,3             ***
                       00032      * ***     2 IAC=65+(IAC-64)/256      ***
                       00033      * ***       GO TO 4                  ***
                       00034      * ***     3 IAC=(IAC-64)/256-63      ***
                       00035      * ***     4 IF(IAC)8,8,9             ***
                       00036      * ***     8 IAC=1                    ***
                       00037      * ***     9 RETURN                   ***
                       00038      * ***       END                      ***
                       00039      * ***                                ***
                       00040      * ***                                ***
                       00041      * *** THE EBCDIC CHARACTERS 00-3F AND**
```

```
                       00042      * *** 80-BF ARE REDUCED BY THESE      ***
                       00043      * *** FORMULAS TO A NEGATIVE NUMBER, **
                       00044      * ***SO THEY ARE SET TO THE VALUE FOR**
                       00045      * ***A BLANK. THE 029 KEYPUNCH CHAR-***
                       00046      * *** ACTER SET LIES IN THE EBCDIC   ***
                       00047      * *** CODE RANGES 40-7F AND C0-FF.   ***
000C                   00048            END
   000 OVERFLOW SECTORS SPECIFIED
   000 OVERFLOW SECTORS REQUIRED
   002 SYMBOLS DEFINED
    NO ERROR(S) AND   NO WARNING(S)  FLAGGED IN ABOVE ASSEMBLY

// DUP
*DELETE             RECNE
CART ID 1234   DB ADDR   21C0    DB CNT    0002
*STORE      WS  UA  RECNE
CART ID 1234   DB ADDR   22CE    DB CNT    0002

// FOR
*ONE WORD INTEGERS
*LIST SOURCE PROGRAM
      SUBROUTINE RECNP(F)
C *** ****************************************************** ***
C *** RECNP(F) CONVERTS THE VALUE OF F INTO A STRING          ***
C *** OF EBCDIC CHARACTERS IN THE FORM BSD. DDDDDESDD         ***
C *** WHERE B IS A BLANK, D IS A DIGIT AND S IS A SIGN, AND   ***
C *** OUTPUTS IT THROUGH RECW. S IS A BLANK WHEN              ***
C *** POSITIVE                                                ***
C *** ****************************************************** ***
      COMMON IAC,ILC,IXL,ILC0
      COMMON IR,IW,IX,IP,IE,IIN,IOUT,IL
      COMMON IPROG(500),IPDL(10,3),ICPL(128),IXEQ(128)
      COMMON IM,KS,KS1,CONST(30)

C ***                                                         ***
C *** INTEGER FUNCTION TO GET THE AL FORMAT                   ***
C *** EQUIVALENT OF THE DIGIT LL                              ***
      IG(LL)=LL*256-4032
C ***                                                         ***
      IF(IW-107)1,1,4
    4 CALL RECW(2)
    1 K=0
      IS=16448
      IAC=16448
      CALL RECW(1)
      V=F
      IF(V)2,7,3
    2 IS=24640
      V=-V
```

```
C ***                                                                 ***
C *** NORMALIZATION                                                   ***
    3 IF(V-10.0)5,6,6
    5 V=V*10.0
      K=K-1
      GO TO 3
    6 V=V*0.1
      K=K+1
      IF(V-10.0)7,6,6
    7 IAC=IS
      CALL RECW(1)
C ***                                                                 ***
C *** ROUNDING AND OUTPUT OF MANTISSA                                 ***
      V=V+5.0E-6
      N=IFIX(V)
      IAC=IG(N)
      CALL RECW(1)
      IAC=19264
      CALL RECW(1)
      DO 8 I=1,5
      V=10.0*(V-FLOAT(N))
      N=IFIX(V)
      IAC=IG(N)
      CALL RECW(1)
    8 CONTINUE
C ***                                                                 ***
C *** OUTPUT OF CHARACTERISTIC                                        ***
      IAC=-15040
      CALL RECW(1)
      IF(K)9,10,10
    9 IAC=24640
      K=-K
      GO TO 11
   10 IAC=16448
   11 CALL RECW(1)
      N=K/10
      IAC=IG(N)
      CALL RECW(1)
      IAC = IG(K-N*10)
      CALL RECW(1)
      RETURN
      END
FEATURES SUPPORTED
 ONE WORD INTEGERS
CORE REQUIREMENTS FOR RECNP
 COMMON     862  VARIABLES       8  PROGRAM     256
RELATIVE ENTRY POINT ADDRESS IS 0028 (HEX)
END OF COMPILATION
// DUP
```

```
*DELETE            RECNP
CART ID 1234   DB ADDR  21C0    DB CNT    0011
*STORE        WS  UA  RECNP
CART ID 1234   DB ADDR  22BF    DB CNT    0011
// FOR
*ONE WORD INTEGERS
*LIST SOURCE PROGRAM
      SUBROUTINE RECQU
C *** *************************************************** ***
C *** RECQU CIMPILES QUOTED STRINGS -- ''XXXXXXX'          ***
C *** *************************************************** ***
      COMMON IAC,ILC,IXL,ILC0
      COMMON IR,IW,IX,IP,IE,IIN,IOUT,IL
      COMMON IPROG(500),IPDL(10,3),ICPL(128),IXEQ(128)
      COMMON IM,KS,KS1,CONST(30)
C ***                                                      ***
C *** ''XXX COMPILES TO-'', N, X, X, X, WHERE N IS THE     ***
C *** NUMBER SYMBOLS IN THE STRING.                        ***
      IPROG(ILC)=-IAC
      ILC=ILC+1
      IX=ILC
      ILC=ILC+1
      IC=0
  198 CALL RECR
      CALL RECW(1)
      IF(IAC-32064)195,196,195
  195 IPROG(ILC)=IAC
      IC=IC+1
      ILC=ILC+1
      IF(ILC-496)198,198,5
  196 IPROG(IX)=IC
      RETURN
    5 IE=-2
      CALL RECW(3)
      RETURN
      END

FEATURES SUPPORTED
 ONE WORD INTEGERS
CORE REQUIREMENTS FOR RECQU
 COMMON     862  VARIABLES      2  PROGRAM      98
RELATIVE ENTRY POINT ADDRESS IS 0008 (HEX)
END OF COMPILATION

// DUP
*DELETE            RECQU
CART ID 1234   DB ADDR  21C0    DB CNT    0007
*STORE        WS  UA  RECQU
CART ID 1234   DB ADDR  22C9    DB CNT    0007
```

```
// FOR
*ONE WORD INTEGERS
*LIST SOURCE PROGRAM
      SUBROUTINE RECR
C *** ***************************************************** ***
C *** RECR DOES ALL NON DISK INPUT THROUGH THE BUF           ***
C *** FER ICARD. TO ACCOMODATE 026 AND 029 KEY PUNCHES       ***
C *** RECR EDITS THE INPUT FOR (,),',=                       ***
C *** ***************************************************** ***
      DIMENSION ICARD(80)
      COMMON IAC,ILC,IXL,ILC0
      COMMON IR,IW,IX,IP,IE,IIN,IOUT,IL
      COMMON IPROG(500),IPDL(10,3),ICPL(128),IXEQ(128)
      COMMON IM,KS,KS1,CONST(30)
      IF(IR-80)2,2,1
    1 READ(IIN,201)ICARD
  201 FORMAT(80A1)
      IR=1
    2 IAC = ICARD(IR)
      IF(IIN-2)9,11,9
C ***                                                        ***
C *** CONVERT PERCENT SIGN TO LEFT PARENTHESIS               ***
   11 IF(IAC-27712)3,4,3
    4 IAC=19776
C ***                                                        ***
C *** LOZENGE (OR LESS THAN) TO RIGHT PARENTHESIS            ***
    3 IF(IAC-19520)5,6,5
    6 IAC = 23872
C ***                                                        ***
C *** AT SIGN TO APOSTROPHE                                  ***
    5 IF(IAC-31808)7,8,7
    8 IAC=32064
C ***                                                        ***
C *** NUMBER SIGN TO EQUAL SIGN
    7 IF(IAC-31552)9,10,9
   10 IAC=32320
    9 IR=IR+1
      RETURN
      END

FEATURES SUPPORTED
 ONE WORD INTEGERS
CORE REQUIREMENTS FOR RECR
 COMMON    862  VARIABLES     82  PROGRAM      94
RELATIVE ENTRY POINT ADDRESS IS 0060 (HEX)
END OF COMPILATION
// DUP
*DELETE              RECR
CART ID 1234    DB ADDR  21C0    DB CNT    0007
```

```
      *STORE      WS  UA  RECR
      CART ID 1234   DB ADDR  22C9   DB CNT   0007
      // FOR
      *ONE WORD INTEGERS
      *LIST SOURCE PROGRAM
            SUBROUTINE RECW(J)
      C *** ****************************************************** ***
      C *** RECW(J) IS THE BUFFERED OUTPUT ROUTINE.                 ***
      C *** J=1  DEPOSIT ONE CHARACTER IN THE OUTPUT BUF-           ***
      C ***      FER ILINE, WRITE OUT THE LINE WHEN THE             ***
      C ***      BUFFER IS FILLED,                                  ***
      C ***  =2  WRITE THE BUFFER IMMEDIATELY, UNLESS               ***
      C ***      EMPTY,                                             ***
      C ***  =3  WRITE ERROR MESSAGE -IE.
      C *** ****************************************************** ***
            DIMENSION ILINE(120),MESS(30)
            COMMON IAC,ILC,IXL,ILC0
            COMMON IR,IW,IX,IP,IE,IIN,IOUT,IL
            COMMON IPROG(500),IPDL(10,3),ICPL(128),IXEQ(128)
            COMMON IM,KS,KS1,CONST(30)
        300 FORMAT(30A2)
        301 FORMAT(80A1)
        302 FORMAT(1X,120A1)
            GO TO (1,5,4),J
      C ***                                                         ***
      C *** ERROR MESSAGE OUTPUT                                    ***
          4 READ(2'-IE)MESS
            WRITE (IOUT,300)MESS
            RETURN
      C ***                                                         ***
      C *** SINGLE. CHARACTER WRITING                               ***
          1 GO TO (12,3),IL
         12 IW=IW+1
            ILINE(IW)=IAC
            IF(IOUT-3)8,7,8
          8 IF(IW-80)3,6,9
          9 WRITE(IOUT,301)(ILINE(I),I=81,IW)
            IW=80
          6 WRITE(IOUT,301)(ILINE(I),I=1,IW)
            IW=0
            GO TO 3
          7 IF(IW-120)3,2,2
      C ***                                                         ***
      C *** LINE DUMP                                               ***
          5 IF(IW)3,3,10
         10 IF(IOUT-3)11,2,11
          2 WRITE(IOUT,302)(ILINE(I),I=1,IW)
            IW=0
            GO TO 3
```

```
   11 WRITE(IOUT,301)(ILINE(I),I=1,IW)
      IW=0
    3 RETURN
      END
FEATURES SUPPORTED
 ONE WORD INTEGERS
CORE REQUIREMENTS FOR RECW
 COMMON     862 VARIABLES     154 PROGRAM     216
RELATIVE ENTRY POINT ADDRESS IS 00AB (HEX)
END OF COMPILATION
// DUP
*DELETE              RECW
CART ID 1234   DB ADDR   21C0   DB CNT   000F
*STORE       WS  UA  RECW
CART ID 1234   DB ADDR   22C1   DB CNT   000F
// FOR
*ONE WORD INTEGERS
*LIST SOURCE PROGRAM
      SUBROUTINE RECXC
C *** **************************************************** ***
C *** RECXC INTERPRETS THE CODE COMPILED BY RECCM.          ***
C *** **************************************************** ***
      DIMENSION IRET(100)
      COMMON IAC,ILC,IXL,ILC0
      COMMON IR,IW,IX,IP,IE,IIN,IOUT,IL
      COMMON IPROG(500),IPDL(10,3),ICPL(128),IXEQ(128)
      COMMON IM,KS,KS1,CONST(30)
C ***    ENTER PROGRAM AT ILC0 + 1 WLTH LOWEST              ***
C *** RECURSION DEPTH                                       ***
  300 FORMAT('1')
      IXL=ILC0+1
      IREC=1
C *** EXIT WHEN ENTRY POINT IN MAIN EXPRESSION IS           ***
C *** REACHED.                                              ***
    1 IF(ILC0-IXL)11,7,11
C ***                                                       ***
C *** IPROG(IXL)                                            ***
C ***    NEGATIVE       SUBROUTINE JUMP (CONSULT IXEQ) (2)  ***
C ***    ZERO           'FALSE RETURN                  (6)  ***
C ***    POSITIVE       UNCONDITIONAL JUMP             (9)  ***
   11 IF(IPROG(IXL))2,6,9
    2 IX=-IPROG(IXL)
      IXL=IXL+1
C ***                                                       ***
C *** SUBROUTINE CATEGORIES --IXEQ                          ***
C ***     LES THAN -500     RECURSIVE SUBROUITE       (17)  ***
C ***     LESS THAN -500    RECURSIVE SUBROUITE       (17)  ***
C ***     EQUAL TO -500     SUBROUITE DECLARED             ***
C ***                       RECURSIVE, BUT NEVER           ***
```

```
C ***                       DEFINED                       (18)  ***
C ***      NEGATIVE         NONRECURSIVE SUBBROUTINE            ***
C ***                       DEFINED BY A REC                    ***
C ***                       EXPRESSION                    (16)  ***
C ***      ZERO             UNDEFINED OPERATOR            (23)  ***
C ***      POSITIVE         SYSTEM SUBROUTINE                   ***
C ***                       DEFINED IN RECXQ               (3)  ***
      IX=IXEQ(IX)
C ***     TEST DATA SWITCH 5, TERMINATE IF ON.                  ***
      CALL RECDS(5,L)
      GO TO (7,4),L
    4 IF(IX)5,23,3
C ***     SYSTEM SUBROUTINES                                    ***
    3 CALL RECXQ
C ***     TEST THE ERROR FLAG. IF IE IS NEGATIVE, RETURN        ***
C ***     TO MONITOR PHASE.                                     ***
      IF(IE)7,1,1
C ***     REC DEFINED SUBROUTINES                               ***
    5 IF(IX+500)17,18,16
C         NONRECURSIVE SUBROUTINES                              ***
   16 IIX=-IX
      IPROG(IIX)=IXL+1
   22 IXL=IIX+1
      GO TO 1
C ***     RECURSIVE SUBROUTINES                                 ***
   17 IIX=-IX-500
      IF(IREC-100)10,10,12
   10 IRET(IREC)=IXL+1
      IREC=IREC+1
      GO TO 22
    6 IXL=IPROG(IXL+1)
      IF(ILC0-IXL)8,7,8
C ***                                                           ***
C *** MAIN PROGRAM RETURN
    7 CALL RECW(2)
      IF(IOUT-3)15,14,15
   14 WRITE(3,300)
   15 RETURN
    8 IF(IPROG(IXL)-2000)21,20,20
C ***     RECURSIVE SUBROUTINE 'FALSE' RETURNS                  ***
   20 IREC=IREC-1
      IXL=IRET(IREC)-1
      GO TO 1
C ***     NONRECURSIVE SUBROUTINE 'FALSE' RETURNS               ***
   21 IXL=IPROG(IXL)-1
      GO TO 1
C ***     UNCONDITIONAL TRASFERS AND 'TRUE' RETURNS             ***
    9 IXL=IPROG(IXL)
      IF(IXL-2000)1,13,13
```

```
C ***     RECURSIVE SUBROUTINE 'TRUE' RETURNS                      ***
   13 IREC=IREC-1
      IXL=IRET(IREC)
      GO TO 1
C ***                                                              ***
C *** ERROR CONDITIONS                                             ***
   23 IE=-13
      GO TO 19
   18 IE=-12
      GO TO 19
   12 IE=-3
   19 CALL RECW(3)
      GO TO 7
      END
```

```
FEATURES SUPPORTED
 ONE WORD INTEGERS
CORE REQUIREMENTS FOR RECXC
 COMMON     862 VARIABLES    104 PROGRAM     282
RELATIVE ENTRY POINT ADDRESS IS 0074 (HEX)
END OF COMPILATION

// DUP
*DELETE             RECXC
CART ID 1234   DB ADDR  21C0   DB CNT  0012
*STORE      WS  UA  RECXC
CART ID 1234   DB ADDR  22BE   DB CNT  0012
// FOR
*ONE WORD INTEGERS
*LIST SOURCE PROGRAM
      SUBROUTINE RECXQ
C *** ****************************************************** ***
C *** RECXQ IMPLEMENTS INTERNAL SUBROUTINES                  ***
C *** ****************************************************** ***
      DIMENSION PDL(200),SAVE(10)
      COMMON IAC,ILC,IXL,ILC0
      COMMON IR,IW,IX,IP,IE,IIN,IOUT,IL
      COMMON IPROG(500),IPDL(10,3),ICPL(128),IXEQ(128)
      COMMON IM,KS,KS1,CONST(30)
      GO TO (1,1,1,1,1,1,1,1,1,1,1,1,1,2,2,2,2,2,2,3,3,3,3,24,25,
     *26,27,28,29,30),IX
C ***                                                              ***
C *** UNARY OPERATIONS                                             ***
C ***                                                              ***
    1 IF(IM-1)90,90,100
C ***                                                              ***
C ***         A   C   E  H M N O Q S 'A 'L 'S 0                    ***
  100 GO TO (101,201,301,4,5,6,7,8,9,10,11,12,13),IX
C ***                                                              ***
```

```
C ***                                                           ***
C *** -------------- ABSOLUTE VALUE                             ***
  101 PDL(IM-1)=ABS(PDL(IM-1))
      RETURN
C ***                                                           ***
C *** -------------- COSINE                                     ***
  201 PDL(IM-1)=COS(PDL(IM-1))
      RETURN
C ***                                                           ***
C *** -------------- EXPONENTIAL                                ***
  301 PDL(IM-1)=EXP(PDL(IM-1))
      RETURN
C ***                                                           ***
C *** -------------- HYPERBOLIC TANGENT                         ***
    4 PDL(IM-1)=TANH(PDL(IM-1))
      RETURN
C ***                                                           ***
C *** -------------- SIGN CHANGE                                ***
    5 PDL(IM-1)=-PDL(IM-1)
      RETURN
C ***                                                           ***
C *** -------------- NEGATIVE TEST                              ***
    6 IF(PDL(IM-1))95,96,96
C ***                                                           ***
C *** -------------- OUTPUT                                     ***
    7 CALL RECNP(PDL(IM-1))
      RETURN
C ***                                                           ***
C *** -------------- SQUARE ROOT                                ***
    8 PDL(IM-1)=SQRT(PDL(IM-1))
      RETURN
C ***                                                           ***
C *** -------------- STORE                                      ***
    9 K=IPROG(IXL)
      SAVE(K)=PDL(IM-1)
      GO TO 95
C ***                                                           ***
C *** -------------- ARC TANGENT                                ***
C******* CHECAR
   10 PDL(IM-1)=ATAN(PDL(IM-1))
      RETURN
C ***                                                           ***
C *** -------------- NATURAL LOGARITHM                          ***
   11 PDL(IM-1)=ALOG(PDL(IM-1))
      RETURN
C ***                                                           ***
C *** -------------- SINE                                       ***
   12 PDL(IM-1)=SIN(PDL(IM-1))
      RETURN
```

```
C ***                                                                    ***
C *** -------------- ZERO TEST                                           ***
   13 IF(ABS(PDL(IM-1))-5.0E-6)95,95,96
C ***                                                                    ***
C *** BINARY OPERATIONS
C ***                                                                    ***
    2 IF(IM-2)90,90,200
  200 IX=IX-13
      IM=IM-1
C ***                                                                    ***
C ***          B   +   -   *   J   /                                     ***
      GO TO (14,15,16,17,18,19),IX
C ***                                                                    ***
C *** -------------- EXPONENTIATION                                      ***
   14 PDL(IM-1)=PDL(IM-1)**PDL(IM)
      RETURN
C ***                                                                    ***
C *** -------------- ADDITION                                            ***
   15 PDL(IM-1)=PDL(IM-1)+PDL(IM)
      RETURN
C ***                                                                    ***
C *** --------------  SUBTRACTION                                        ***
   16 PDL(IM-1)=PDL(IM-1)-PDL(IM)
      RETURN
C ***                                                                    ***
C *** -------------- MULTIPLICATION                                      ***
   17 PDL(IM-1)=PDL(IM-1)*PDL(IM)
      RETURN
C ***                                                                    ***
C *** -------------- EQUAL TEST                                          ***
   18 IM=IM+1
      IF(ABS(PDL(IM-1)-PDL(IM-2))-5.0E-6)95,95,96
C ***                                                                    ***
C *** -------------- DIVISION                                            ***
   19 PDL(IM-1)=PDL(IM-1)/PDL(IM)
      RETURN
C ***                                                                    ***
C *** DATA GENERATORS                                                    ***
C ***                                                                    ***
    3 IF(IM-200)300,300,91
  300 IM=IM+1
      IX=IX-19
```

```
C ***                                                                  ***
C ***         '/ F I P                                                 ***
      GO TO (20,21,22,23),IX
C ***                                                                  ***
C ***                                                                  ***
C *** -------------- PROGRAM CONSTANT                                  ***
   20 K=IPROG(IXL)
      PDL(IM-1)=CONST(K)
      GO TO 95
C ***                                                                  ***
C *** -------------- FETCH                                             ***
   21 K=IPROG(IXL)
      PDL(IM-1)=SAVE(K)
      GO TO 95
C ***                                                                  ***
C *** -------------- INPUT                                             ***
   22 CALL RECR
      IF(IAC-16448)320,22,320
  320 IF(IAC-32064)92,324,92
  324 CALL RECR
      IF(IAC-24896)92,321,92
  321 CALL RECNC(0,PDL(IM-1),I)
  327 IF(IAC-16448)322,326,322
  322 IF(IAC-32064)92,96,92
  326 CALL RECR
      GO TO 327
C ***                                                                  ***
C *** -------------- PUSH                                              ***
   23 IF(IM-2)90,90,323
  323 PDL(IM-1)=PDL(IM-2)
      RETURN
C ***                                                                  ***
C *** R,W,&,=X,X,$ AND L                                               ***
C ***                                                                  ***
C ***                                                                  ***
C *** -------------- READ                                              ***
   24 CALL RECR
      RETURN
C ***                                                                  ***
C *** -------------- WRITE                                             ***
   25 CALL RECW(1)
      RETURN
C ***                                                                  ***
C *** -------------- OUTPUT QUOTED STRINGS
   26 IC=IPROG(IXL)
      IF(IC)95,95,261
  261 DO 262 I=1,IC
      IXL=IXL+1
      IAC=IPROG(IXL)
```

```
      CALL RECW(1)
  262 CONTINUE
      GO TO 95
C ***                                                              ***
C *** --------------- CHARACTER EQUALITY TEST                      ***
   27 IF(IAC-IPROG(IXL))95,270,95
  270 IXL=IXL+1
   95 IXL=IXL+1
   96 RETURN
C ***                                                              ***
C *** --------------- OUTPUT WRITE BUFFER                          ***
   28 CALL RECW(2)
      RETURN
C ***                                                              ***
C *** --------------- COUNTERS                                     ***
   29 CALL RECCN
      RETURN
C ***                                                              ***
C *** --------------- LIFT                                         ***
   30 IF(IM-1)96,96,302
  302 IM=IM-1
      GO TO 96
C ***                                                              ***
C *** --------------- ERRORS DETECTED                              ***
   90 IE=-4
      GO TO 93
   91 IE=-5
      GO TO 93
   92 IE=-11
   93 CALL RECW(3)
      RETURN
      END
FEATURES SUPPORTED
 ONE WORD INTEGERS
CORE REQUIREMENTS FOR RECXQ
 COMMON     862 VARIABLES    428 PROGRAM    822
RELATIVE ENTRY POINT ADDRESS IS 01BB (HEX)
END OF COMPILATION

// DUP
*DELETE            RECXQ
CART ID 1234   DB ADDR  21C0   DB CNT   0035
*STORE      WS  UA  RECXQ
CART ID 1234   DB ADDR  229B   DB CNT   0035

// FOR
*NAME RECDO
*IOCS(CARD,1132 PRINTER,DISK)
*ONE WORD INTEGERS
```

```
*LIST SOURCE PROGRAM
C *** ****************************************************** ***
C *** RECDO READS, STORES ON DISK, READS BACK AND             ***
C *** PRINTS CHARACTER DEFINITIONS FOR MONITOR, COM-          ***
C *** PILATION AND EXECUTION CODES                            ***
C *** ****************************************************** ***
      DIMENSION ICARD(64),JCARD(128),KCARD(128),LCARD(128)
      DIMENSION ICRD1(32),ICRD2(32)
      DIMENSION JCRD1(32),JCRD2(32),JCRD3(32),JCRD4(32)
      DIMENSION KCRD1(32),KCRD2(32),KCRD3(32),KCRD4(32)
      DIMENSION LCRD1(32),LCRD2(32),LCRD3(32),LCRD4(32)
      EQUIVALENCE (ICARD( 1), ICRD1( 1)),
     *            (ICARD(33), ICRD2( 1))
      EQUIVALENCE (JCARD( 1), JCRD1( 1)),
     *            (JCARD(33), JCRD2( 1))
      EQUIVALENCE (JCARD(65), JCRD3( 1)),
     *            (JCARD(97), JCRD4( 1))
      EQUIVALENCE (KCARD(65), KCRD3( 1)),
     *            (KCARD(97), KCRD4( 1))
      EQUIVALENCE (KCARD( 1), KCRD1( 1)),
     *            (KCARD(33), KCRD2( 1))
      EQUIVALENCE (LCARD( 1), LCRD1( 1)),
     *            (LCARD(33), LCRD2( 1))
      EQUIVALENCE (LCARD(65), LCRD3( 1)),
     *            (LCARD(97), LCRD4( 1))
      DEFINE FILE 1 (3,128,U,KF)
  100 FORMAT(32(2X,A1))
  101 FORMAT(32I3)
  102 FORMAT(/)
  200 FORMAT(64A1)
  201 FORMAT(3(32I2/),32I2)
      READ(2,200)ICARD
      READ(2,201)JCARD
      READ(2,201)KCARD
      READ(2,201)LCARD
      WRITE(1'1)JCARD
      WRITE(1'2)KCARD
      WRITE(1'3)LCARD
      READ(1'1)JCARD
      READ(1'2)KCARD
      READ(1'3)LCARD
      WRITE(3,100)ICRD1
      WRITE(3,101)JCRD1,KCRD1,LCRD1
      WRITE(3,102)
      WRITE(3,100)ICRD2
      WRITE(3,101)JCRD2,KCRD2,LCRD2
      WRITE(3,102)
      WRITE(3,100)ICRD1
      WRITE(3,101)JCRD3,KCRD3,LCRD3
```

```
      WRITE(3,102)
      WRITE(3,100)ICRD2
      WRITE(3,101)JCRD4,KCRD4,LCRD4
      CALL EXIT
      END
FEATURES SUPPORTED
 ONE WORD INTEGERS
 IOCS
CORE REQUIREMENTS FOR RECDO
 COMMON      0 VARIABLES    458 PROGRAM    206
END OF COMPILATION
// DUP
*STORE      WS  UA  RECDO
CART ID 1234   DB ADDR  225D   DB CNT   0011
*STOREDATA WS  UA  RECAT 0002
CART ID 1234   DB ADDR  2270   DB CNT   0020

// FOR
*NAME RECER
*LIST SOURCE PROGRAM
*IOCS(CARD,1132 PRINTER,DISK)
*ONE WORD INTEGERS
C *** ***************************************************** ***
C *** RECER READS. STORES ON DISK,READS BACK AND            ***
C *** PRINTS 20 ERROR MESSAGES                              ***
C *** ***************************************************** ***
      DIMENSION MESS(30)
      DEFINE FILE 2 (20,30,U,KF)
  200 FORMAT(30A2)
  201 FORMAT(1X,30A2)
      DO 10 I=1,20
      READ(2,200)MESS
      WRITE(2'I)MESS
      READ(2'I)MESS
      WRITE(3,201)MESS
   10 CONTINUE
      CALL EXIT
      END
FEATURES SUPPORTED
 ONE WORD INTEGERS
 IOCS
CORE REQUIREMENTS FOR RECER
 COMMON      0 VARIABLES     40 PROGRAM     76
END OF COMPILATION
// DUP
*STORE      WS  UA  RECER
CART ID 1234   DB ADDR  2290   DB CNT   0007
*STOREDATA WS  UA  RECEM 0002
CART ID 1234   DB ADDR  22A0   DB CNT   0020
```

```
// JOB REC
LOG DRIVE    CART SPEC    CART AVAIL   PHY DRIVE
  0000         1234         1234         0000
V2 M12   ACTUAL 16K   CONFIG 16K
// * 690017 ESCUELA SUPERIOR DE FISICA Y MATEMATICAS
// XEQ REC    L   2
*LOCALREC,RECCA,RECCC,RECCO,RECKO,RECMO,RECNP,RECQU
*FILES(1,RECAT),(2,RECEM)
FILES ALLOCATION
    1 0218  0002  1234 RECAT
    2 021A  0002  1234 RECEM
STORAGE ALLOCATION
R 41  1BF4 (HEX) WDS UNUSED BY CORE LOAD
CALL TRANSFER VECTOR
 FLN    1BF8
 FXPN   1A84
 RECCN  1D1C
 FAXBX  1C7E
 FSIN   1A14
 FALOG  1BF6
 FATAN  1B74
 FSQRT  1B3A
 FTANH  1AF8
 FEXP   1A82
 FCOS   1A0C
 FABS   1A00
 DATSW  19DE
 RECXQ  15CB
 RECDS  13ED
 RECFC  122E
 RECNC  10BE
 RECNE  1098
 RECW   0FD1
 RECR   0ED6
 RECXC  0D2A
 RECCM  0C17
 RECQU  1E22 LOCAL
 RECNP  1E42 LOCAL
 RECMO  1E71 LOCAL
 RECKO  1E24 LOCAL
 RECCO  1E2E LOCAL
 RECCC  1E25 LOCAL
 RECCA  1E24 LOCAL
LIBF TRANSFER VECTOR
```

```
     FDIV   1CB8
     FGETP  1D74
     FDVR   1D5E
     FDIVX  1CB4
     FMPYX  12C0
     FSUBX  125F
     FLDX   0BF0
     SIOI   0634
     HOLTB  19DB
     FADDX  126B
     XMDS   198C
     FARC   196A
     NORM   1940
     FAXI   18F6
     SIOAI  063A
     SRED   052D
     EBCTB  13D7
     HOLEZ  1368
     GETAD  1358
     PAUSE  1342
     FSBR   132E
     FLOAT  1320
     IFIX   12F4
     FADD   1265
     FMPY   12C4
     FSUB   125A
     SNR    1252
     SIOIX  05BF
     FSTOX  0BD4
     SUBSC  0E58
     SUBIN  0E38
     SDAI   02AA
     SDRED  02CC
     SCOMP  060C
     SWRT   0528
     FSTO   0BD8
     FLD    0BF4
     PRNTZ  0AEC
     CARDZ  0A3C
     TYPEZ  09D2
     SFIO   0649
     SDFIO  0330
SYSTEM SUBROUTINES
     ILS04  00C4
     ILS02  00B3
     ILS01  1FDE
     ILS00  1FF7
     FLIPR  1DB4
         0219 (HEX) IS THE EXECUTION ADDR
```

```
C DAMPED OSILLATIONS (Y=SIN(3*X)*EXP(-0.3*X)
* ('/ 'S1L(F1O'/3'*'SF1'/-0.3'*E*O'/1.0'&(N''*','' ''/-0.04'&.)LXF1'/0.15'&S1L
$50$.,),)
  0.00000E 00  0.00000E 00                          *
  1.50000E-01  4.15826E-01                                     *
  3.00000E-01  7.15906E-01                                            *
  4.50000E-01  8.52504E-01                                                *
  6.00000E-01  8.13425E-01                                               *
  7.50000E-01  6.21304E-01                                         *
  9.00000E-01  3.26253E-01                                   *
  1.05000E 00 -6.13488E-03                         *
  1.20000E 00 -3.08735E-01                   *
  1.35000E 00 -5.25927E-01              *
  1.50000E 00 -6.23300E-01            *
  1.65000E 00 -5.92444E-01             *
  1.80000E 00 -4.50328E-01                *
  1.95000E 00 -2.33855E-01                    *
  2.10000E 00  8.95249E-03                         *
  2.25000E 00  2.29140E-01                               *
  2.40000E 00  3.86318E-01                                  *
  2.55000E 00  4.55686E-01                                    *
  2.70000E 00  4.31465E-01                                   *
  2.85000E 00  3.26366E-01                                *
  3.00000E 00  1.67559E-01                            *
  3.15000E 00 -9.79736E-03                         *
  3.29999E 00 -1.70005E-01                     *
  3.44999E 00 -2.83736E-01                  *
  3.59999E 00 -3.33121E-01                 *
  3.74999E 00 -3.14203E-01                 *
  3.89999E 00 -2.36499E-01                   *
  4.04999E 00 -1.20008E-01                       *
  4.19999E 00  9.53147E-03                         *
  4.34999E 00  1.26089E-01                             *
  4.49999E 00  2.08370E-01                               *
  4.64999E 00  2.43504E-01                                *
  4.79999E 00  2.28793E-01                               *
  4.94999E 00  1.71356E-01                            *
  5.09999E 00  8.59142E-02                        *
  5.24999E 00 -8.69345E-03                         *
  5.39999E 00 -9.34868E-02                      *
  5.54999E 00 -1.53006E-01                     *
  5.69999E 00 -1.77983E-01                    *
  5.84999E 00 -1.66588E-01                    *
  5.99999E 00 -1.24142E-01                      *
  6.14999E 00 -6.14790E-02                        *
  6.29999E 00  7.61194E-03                         *
  6.44999E 00  6.92922E-02                           *
  6.59998E 00  1.12341E-01                            *
  6.74999E 00  1.30083E-01                             *
  6.89999E 00  1.21286E-01                            *
  7.04998E 00  8.99251E-02                         *
  7.19998E 00  4.39726E-02                        *
  7.34998E 00 -6.47955E-03                        *
  7.49998E 00 -5.13438E-02                      *
```

```
C THE EXPRESSION F(X,Y)=(X*X&Y*Y)**5-(8*(X*X-Y*Y)*X*Y)**2
C IS EVALUATED 50*74 TIMES. THE CONDITION F(X,Y)=0 IS THE
C BOUNDARY CURVE R = 2*SIN(4*THETA). AN * IS PUT OUT IF
C F(X,Y) IS NEGATIVE AND A BLANK IF IT IS POSITIVE
* ('/-2'S0L($50$'/-2'S1L($74$F1P*F0P*&PPPP****F1P*F0P*-
F1*F0*'/8'*P*-(N''*','' ',)LF1'/0.054'&S1L.
,)XF0'/0.08'&S0L.,),)
```

```
                              ***                        ***
                            ******                      ******
                           *******                      *******
                          *********                    *********
                          **********                   **********
                          **********                   **********
                          **********                   **********
                          **********                   **********
                           **********                  **********
                           **********                  **********
                            **********                **********
                             *********                *********
                             **********              **********
     **********              **********              **********              **********
   ****************          *********                *********            ****************
  ***************** *****    *********                *******         ******************
  ********************       *******                  ********        *********************
   ********************      *******                  *******         **********************
    ********************     ******                   ******          **********************
     ********************   *****                     *****           ********************
      *********************                                           *********************
       ************************                      ***********************
        ****************** ** **                  ***** **********************
                             ********************************
                                       *
                              ******************************
           ****************** ** **                ** ** ****************
          ***********************                  ***********************
         ********************    *****              *****    *********************
        ********************      ******            ******    ********************
       *********************       *******          *******     ********************
      ********************          *******        *******        ********************
     ******************              ********      ********        ********************
    ******************                ********      *******         *******************
   *****************                  *********    *********         ****************
    **********                         **********  **********          **********
                                        **********  **********
                                        **********  **********
                                        **********  **********
                                        **********  **********
                                        **********  **********
                                         *********  **********
                                         **********  *********
                                         **********  *********
                                         *********    *********
                                         *********    *********
                                          ********    ********
                                          *******     *******
                                           ******     ******
                                            ***        ***
```

```
 C FACTORIAL DE NUMERO
* N'R
(N,0L'/1',P'/1'-'*RECURSION' 'R*,)'RL
      0    -22      5     20    -49     11    -20    -98      1     20    -24
    -98      2    -33    -90     19    -29     20      0      1
('/ 'S0L($10$F0'/1'&S0 O 'R O X.,),) L
 1.00000E 00  1.00000E 00
 2.00000E 00  2.00000E 00
 3.00000E 00  6.00000E 00
 4.00000E 00  2.40000E 01
 5.00000E 00  1.20000E 02
 6.00000E 00  7.20000E 02
 7.00000E 00  5.04000E 03
 8.00000E 00  4.03200E 04
 9.00000E 00  3.62880E 05
 1.00000E 01  3.62880E 06

C SIMPSON INTEGRATION
C   4*I(0,1)(1/1&X**2)
*                                                                              (
F6 P F1 P * & / ,)Y
(IOIOXS1-IOXS3'/2'*'/S4'/3'/S5L'/1'S6'/'
(F3F6-S3N,LY&F1F4&S1LY'/4'*&F1F4&S1LY&X.
)LF5*,)'R
('R'/4'*''PI='OX,)
 1.00000E 00  0.00000E 00
 4.00000E 01
PI=  3.14157E 00

C THE EXPRESSION F(X,Y)=(X*X&Y*Y)**5-(8*(X*X-Y*Y)*X*Y)**2
C IS EVALUATED 50*74 TIMES. THE CONDITION F(X,Y)=0 IS THE
C BOUNDARY CURVE R = 2*SIN(4*THETA). AN * IS PUT OUT IF
C F(X,Y) IS NEGATIVE AND A BLANK IF IT IS POSITIVE
* ( '/-2'S0L($50$'/-2'S1L($74$F1P*F0P*&PPPP****F1P*F0P*-
F1*F0*'/8'*P*-(N''*','' ',)LF1'/0.054'&S1L.
,)XF0'/0.08'&S0L.,),)
```